\documentclass[journal]{IEEEtran}
\usepackage[T1]{fontenc}
\usepackage{textcomp}
\usepackage[latin9]{inputenc}
\usepackage{color}
\definecolor{document_fontcolor}{rgb}{0.210938, 0.355469, 0.332031}
\color{document_fontcolor}
\usepackage{array}
\usepackage{units}
\usepackage{url}
\usepackage{multirow}
\usepackage{amsmath}
\usepackage{amssymb}
\usepackage{graphicx}
\usepackage[bookmarks=true,bookmarksnumbered=true,bookmarksopen=true,bookmarksopenlevel=1,
 breaklinks=false,pdfborder={0 0 0},pdfborderstyle={},backref=false,colorlinks=false]
 {hyperref}
\hypersetup{pdftitle={Your Title},
 pdfauthor={Your Name},
 pdfpagelayout=OneColumn, pdfnewwindow=true, pdfstartview=XYZ, plainpages=false}

\makeatletter

\newcommand{\noun}[1]{\textsc{#1}}
\providecommand{\tabularnewline}{\\}

\IEEEoverridecommandlockouts
\usepackage[caption=false,font=footnotesize]{subfig}
\usepackage{cite}
\usepackage{color}
\usepackage{graphicx,import}

\makeatother

\begin{document}
\global\long\def\quat#1{\boldsymbol{#1}}%

\global\long\def\dq#1{\underline{\boldsymbol{#1}}}%

\global\long\def\hp{\mathbb{H}_{p}}%

\global\long\def\dotmul#1#2{\left\langle #1,#2\right\rangle }%

\global\long\def\partialfrac#1#2{\frac{\partial\left(#1\right)}{\partial#2}}%

\global\long\def\totalderivative#1#2{\frac{d}{d#2}\left(#1\right)}%

\global\long\def\mymatrix#1{\boldsymbol{#1}}%

\global\long\def\vecthree#1{\operatorname{v}_{3}\left(#1\right)}%

\global\long\def\vecfour#1{\operatorname{v}_{4}\left(#1\right)}%

\global\long\def\haminuseight#1{\overset{-}{\mymatrix H}_{8}\left(#1\right)}%

\global\long\def\hapluseight#1{\overset{+}{\mymatrix H}_{8}\left(#1\right)}%

\global\long\def\haminus#1{\overset{-}{\mymatrix H}_{4}\left(#1\right)}%

\global\long\def\haplus#1{\overset{+}{\mymatrix H}_{4}\left(#1\right)}%

\global\long\def\norm#1{\left\Vert #1\right\Vert }%

\global\long\def\abs#1{\left|#1\right|}%

\global\long\def\conj#1{#1^{*}}%

\global\long\def\veceight#1{\operatorname{v}_{8}\left(#1\right)}%

\global\long\def\myvec#1{\boldsymbol{#1}}%

\global\long\def\imi{\hat{\imath}}%

\global\long\def\imj{\hat{\jmath}}%

\global\long\def\imk{\hat{k}}%

\global\long\def\bbloss{L_{\text{bb}}}%

\global\long\def\locloss{L_{\text{loc}}}%

\global\long\def\confloss{L_{\text{conf}}}%

\global\long\def\bouloss{L_{\text{b}}}%

\global\long\def\diceloss{L_{\text{dice}}}%

\global\long\def\javecc{\boldsymbol{J}_{c}^{\lg}}%

\global\long\def\vece{\quat p_{e}^{\lg}}%

\global\long\def\dotvece{\dot{\quat p}_{e}^{\lg}}%

\global\long\def\veca{\quat p_{\la}^{\lg}}%

\global\long\def\dotveca{\dot{\quat p}_{\la}^{\lg}}%

\global\long\def\javeca{\boldsymbol{J}_{\la}^{\lg}}%

\global\long\def\vecasi{\quat p_{\si}^{\la}}%

\global\long\def\dotvecasi{\dot{\quat p}_{\si}^{\la}}%

\global\long\def\javecasi{\boldsymbol{J}_{\si}^{\la}}%

\global\long\def\vecsi{\quat t_{\si}^{\lg}}%

\global\long\def\dotvecsi{\dot{\quat t}_{\si}^{\lg}}%

\global\long\def\javecsi{\boldsymbol{J}_{\si}^{\lg}}%

\global\long\def\fworld{\mathcal{F}_{\text{W}}}%

\global\long\def\flg{\mathcal{F}_{\lg}}%

\global\long\def\fdrill{\mathcal{F}_{\text{R}}}%

\title{Autonomous Robotic Drilling System for Mice Cranial Window Creation}
\author{Enduo~Zhao, Murilo~M.~Marinho,~\IEEEmembership{Senior Member,~IEEE,}
and Kanako~Harada,~\IEEEmembership{Member,~IEEE}\thanks{This work
was supported by JST Moonshot R\&D JPMJMS2033.} \thanks{(\emph{Corresponding
author:} Murilo~M.~Marinho)}\thanks{Enduo~Zhao was with the Department
of Mechanical Engineering, Graduate School of Engineering, The University
of Tokyo, Tokyo, Japan, when this work was conducted. He is currently
with the School of Biomedical Engineering, Tsinghua University, Beijing,
China. \texttt{Email:endowzhao@mail.tsinghua.edu.cn.} }\thanks{Murilo
M. Marinho is a visiting researcher with the Graduate School of Medicine,
the University of Tokyo, Tokyo, Japan. He is also with the Department
of Electrical and Electronic Engineering, the University of Manchester,
Manchester, UK. He has been supported by the Robotics and AI Collaboration
(RAICo). \texttt{Email:murilo.marinho@manchester.ac.uk.}}\thanks{Kanako~Harada
is with the Center for Disease Biology and Integrative Medicine, Graduate
School of Medicine, The University of Tokyo, Tokyo, Japan. \texttt{Email:kanakoharada@g.ecc.u-tokyo.ac.jp}.}}
\maketitle
\begin{abstract}
Robotic assistance for experimental manipulation in the life sciences
is expected to enable favorable outcomes, regardless of the skill
of the scientist. Experimental specimens in the life sciences are
subject to individual variability and hence require intricate algorithms
for successful autonomous robotic control. As a use case, we are studying
the cranial window creation in mice. This operation requires the removal
of an 8-mm circular patch of the skull, which is approximately 300
\textit{\textmu }m thick, but the shape and thickness of the mouse
skull significantly varies depending on the strain of the mouse, sex,
and age. In this work, we develop an autonomous robotic drilling system
with no offline planning, consisting of a trajectory planner with
execution-time feedback with drilling completion level recognition
based on image and force information. In the experiments, we first
evaluate the image-and-force-based drilling completion level recognition
by comparing it with other state-of-the-art deep learning image processing
methods and conduct an ablation study in eggshell drilling to evaluate
the impact of each module on system performance. Finally, the system
performance is further evaluated in postmortem mice, achieving a success
rate of 70\% (14/20 trials) with an average drilling time of 9.3 min. 

\textit{Note to Practitioners}\textemdash This paper addresses the
challenge of drilling over a trajectory with specified depth using
image and force information. The proposed strategy compensates at
execution time for unknown characteristics such as shape, thickness,
and operational noise which are common to organic matter such as eggs
and mice skulls. The trajectory is general but for this work, it is
evaluated as circular. Pre-operatively, sufficient training data (e.g.,
videos) must be obtained and manually annotated with the pixel-wise
completion level. At the training stage, this annotation for completion
can be discretized into classes (e.g., 0\%, 25\%, 50\%, 75\%, 100\%)
to facilitate annotation. These annotations are utilized to train
the entire multimodal model, including image and force, therefore
no additional force annotation is needed. This single data gathering
generalizes over individual differences because the visual features
that correlate with pixel-wise completion are not dependent on the
overall shape and size of the surface. We apply this strategy to many
fragile targets with individual variations such as raw chicken eggs
and postmortem mice skulls. Nonetheless, a success rate of 100\% has
not yet been achieved because even human annotators struggle to differentiate
completion levels in mice skulls. The inclusion of other sensorial
modalities might be needed for further progress.
\end{abstract}


\section{Introduction}

\begin{figure}[tbh]
\begin{centering}
\includegraphics[width=1\columnwidth]{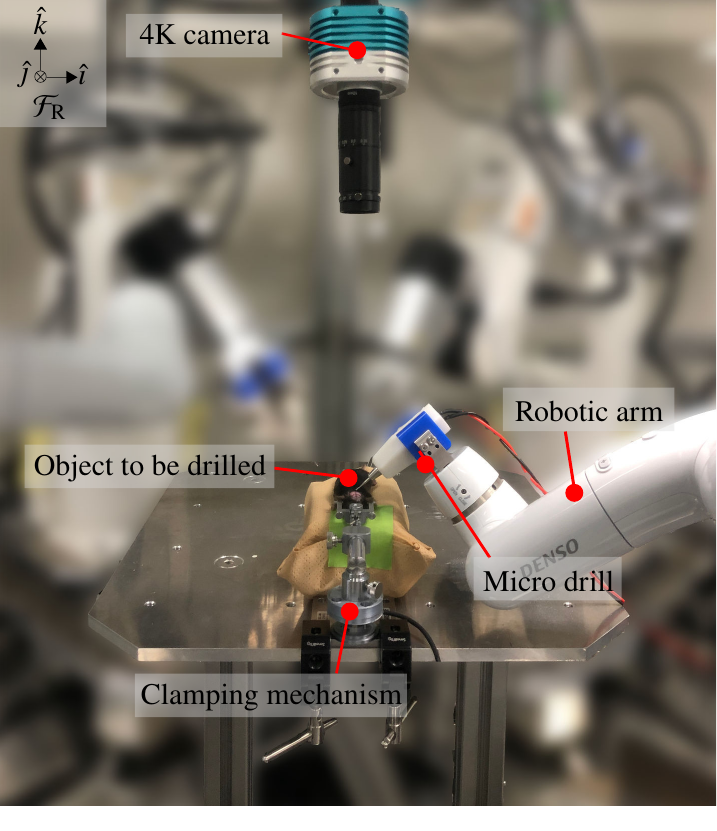}
\par\end{centering}

\caption{\label{fig:robot_system}The system setup used in this work consists
of one of the 8-degrees-of-freedom robotic branches of an AI-robot
platform for scientific exploration \cite{Marinho2024}. For this
work, we attached a micro drill as an end effector. In addition, we
use a 4K camera and a clamping mechanism for the object being drilled.}
\end{figure}

\IEEEPARstart{A}{} cranial window is a transparent observation window
carefully created in the skull of a mouse, providing direct visualization
and access to the brain for experimental purposes. For example, a
cranial window can be utilized to observe human organoids that are
implanted into the mouse brain \cite{Koike2019}. Cranial window creation
involves using a micro drill under a microscope to extract an 8-mm
circular patch of the mouse skull, which is then replaced with a cover
glass \cite{Holtmaat2009,Goldey2014}. This technique is crucial for
advancing our understanding of human cell growth and developing strategies
for diagnosing, treating, and preventing neurological disorders. However,
variabilities in surface flatness, skull thickness and density complicate
the operation, and the small size and delicate nature of mouse skulls
increase the risk of damaging critical structures during the procedure
\cite{Drew2010,Holtmaat2009,Yang2012}, leading to inflammation, infection,
bleeding, or even the death of the mouse. This constitutes a catastrophic
failure, as it prevents the subsequent implantation and observation
of human cells \cite{Holtmaat2009,Kim2016}. These factors further
compound the complexity and difficulty of the experiment.

In many biological procedures analogous to cranial window creation,
robotic systems equipped with multimodal sensing and real-time control
play a significant role by adaptively performing complex tasks amid
anatomical variability. For instance, Liu \emph{et al.} \cite{Liu2024}
developed an electromagnetic system achieving sub-40\,\textit{\textmu }m
trajectory precision in intraocular microsurgery through disturbance-rejection
control and thermal-optimized actuation, while Wang \emph{et al.}
\cite{Wang2024} developed a multimodal sensor-enabled soft robotic
system that achieves 100\% recognition accuracy even for geometrically
identical objects through feature-level sensor fusion and adaptive
grasping control. Building upon such advancements, robotic systems
are increasingly regarded as promising solutions to overcome many
of the challenges associated with cranial window creation in mice,
by performing repetitive tasks with higher precision, reducing the
task time, and enhancing the quality of the experiments. Numerous
devices and robotic systems have been devised to aid in cranial window
creation. Pohl \emph{et al. }\cite{Pohl2011} designed a robotic module
for assisting mouse craniotomy by monitoring force and sound signals,
estimating drilling penetration, and controlling drill feed speed.
However, further automating cranial window creation using a robotic
system presents greater challenges in terms of the robot's perceptual
and adaptive capabilities. This is mainly due to the uncertainties
in both the external and internal characteristics of the mouse skull,
such as uneven surfaces, non-uniform thickness, fragile membrane beneath
the skull, and potential deformations occurring during drilling. Therefore,
the system needs to be able to obtain and process not only global
information, for global planning of the drilling trajectory, but also
contact signals, for real-time feedback to handle dynamic changes.
Our team is investigating scientific experiments' automation using
a robotic manipulator retrofitted with a micro drill, as shown in
Fig.~\ref{fig:robot_system}, with cranial window creation being
one of the intended automation targets.

As an end goal, cranial window creation must be conducted on anesthetized
mice given the nature of the experiments. However, relying on live
mice for studying autonomous robot control in relatively early stages
of research raises ethical concerns. Therefore, initially we rely
on training procedures used for human technical specialists \cite{andreoli2018egg}
and surgical trainees \cite{okuda2010training}, which comprises removing
a circular patch of chicken eggshell without damaging the underlying
membrane.

In previous research \cite{Zhao2023}, an autonomous robotic drilling
system was developed and preliminarily validated in autonomously drilling
eggshells using image feedback. In this study, by incorporating force
feedback and a plane-fitting algorithm, we report improved efficacy
and speed in eggshell drilling.. With these sizable improvements over
the state-of-the-art, in this work, we show the first-in-the-World
example of effective mouse drilling without pre-processing on \emph{postmortem}
mice, which is a large step towards the automation of the cranial
window creation in live mice. Note that no mice were euthanized in
this research\footnote{The \emph{postmortem} mice used in this work were donated to us after
an unrelated and ethically approved experimentation in a neighboring
research laboratory. The mice would have otherwise been discarded.
Hence, no additional ethical committee approval was sought for the
purposes of this work.}.

Although we showcase the work in a scientific exploration background,
our system can be seen as a multimodal adaptive autonomous robotic
drilling system. The system can generalize to unknown surface and
thickness of the drilled object without additional data thanks to
the trajectory planner and the exteroceptive drilling completion level
recognition.

\begin{table*}[tbh]
\begin{centering}
\caption{\label{tab:related work}Capabilities of the proposed work in contrast
with existing literature.}
\par\end{centering}
\begin{centering}
\begin{tabular}{|c|c||c|c|c|c|c|c|c|}
\hline 
 & This work & \cite{Zhao2023} & \cite{Pak2015} & \emph{\noun{\cite{Ghanbari2019}}} & \cite{Jeong2013} & \cite{hasegawa2023} & \cite{Jia2023} & \cite{Navabi25}\tabularnewline
\hline 
\hline 
Contact signals\textsuperscript{1} & $\surd$ & $\times$ & $\surd$ & $\surd$ & $\times$ & $\surd$ & $\surd$ & $\times$\tabularnewline
\hline 
Global information\textsuperscript{2} & $\surd$ & $\surd$ & $\surd$ & $\surd$ & $\surd$ & $\times$ & $\surd$ & $\surd$\tabularnewline
\hline 
No pre-operative information & $\surd$ & $\surd$ & $\times$ & $\times$ & $\times$ & $\times$ & $\surd$ & $\times$\tabularnewline
\hline 
No accurate simulation required\textsuperscript{3} & $\surd$ & $\surd$ & $\surd$ & $\surd$ & $\surd$ & $\surd$ & $\times$ & $\surd$\tabularnewline
\hline 
No use of specialized sensors\textsuperscript{4} & $\surd$ & $\surd$ & $\times$ & $\surd$ & $\times$ & $\surd$ & $\surd$ & $\times$\tabularnewline
\hline 
Execution-time feedback & $\surd$ & $\surd$ & $\times$ & $\times$ & $\surd$ & $\surd$ & $\surd$ & $\times$\tabularnewline
\hline 
Validated on mice & $\surd$ & $\times$ & $\surd$ & $\surd$ & $\surd$ & $\times$ & $\times$ & $\surd$\tabularnewline
\hline 
Assessment metrics available\textsuperscript{5} & $\surd$ & $\surd$ & $\Delta$ & $\Delta$ & $\Delta$ & $\Delta$ & $\times$ & $\Delta$\tabularnewline
\hline 
\end{tabular}
\par\end{centering}
\begin{raggedright}
\textsuperscript{1}Sensors are used for contact detection/estimation
include force sensor, audio, and electrical conductance.
\par\end{raggedright}
\textsuperscript{2}Sensors are used to collect global information
on the entire surface. These include a camera, force sensors to measure
points in the surface, micro-CT scanner, OCT, and ultrashort pulsed
Ti-sapphire oscillator.

\textsuperscript{3}An accurate computer simulation was part of the
workflow. This means that it would not be applicable to mice given
the complexities of simulating mice's skin, bones, and other structures.

\textsuperscript{4}Specialized sensors were included in the system.
This means the work is more difficult to reproduce and implement broadly.

\textsuperscript{5}Repeat experiments were conducted and general
metrics (success rate, average operation time) are available to assess
the performance of the system. $\Delta$ refers to when only success
rate or speed was evaluated.
\end{table*}

\subsection{Related works}

Some studies have partially addressed the challenges of automating
cranial window creation using robotic systems. Pak \emph{et al. }\cite{Pak2015}
demonstrated a penetration detection strategy using a measurement
circuit that detects electrical conductance between the drill and
the mouse, indicating skull penetration by a sudden increase in conductance.
Hasegawa \textit{et al.} \cite{hasegawa2023} proposed a drilling
system that judges the local penetration by extracting feature values
based on force and sound signals, achieving automatic drilling and
being evaluated on the egg model. These studies based on the perception
of contact signals such as force, sound, and electrical signals can
only make judgments about whether penetration occurs at the local
position in contact with the drill bit, as the perception signals
are generated only during contact. This will result in a lack of global
awareness of the entire drilling trajectory.

To perceive global information, Ghanbari \textit{et al.}\emph{ }\emph{\noun{\cite{Ghanbari2019}}}
developed a cranial microsurgery platform ``Craniobot'' that employs
a surface profiler and micro-computed tomography to assess skull surface
topology and thickness. These data are utilized to create a 3D drilling
path towards the desired depth. Jeong \textit{et al.} \cite{Jeong2013}
utilized the metrology via second harmonic generation to map mouse
skull surfaces and establish a drilling path, then employed plasma-mediated
laser ablation for drilling purposes. Navabi \textit{et al.} \cite{Navabi25}
combined optical coherence tomography (OCT) and machine learning to
obtain the skull anatomy, which was then used for trajectory planning
and automated drilling. These works are capable of the perception
of global information, but obtaining the pre-operative information
of every mouse would be time-consuming due to the individual differences,
and over-reliance on pre-operative information may result in the system\textquoteright s
inadequate execution-time feedback on dynamic changes in the characteristics
of the skull during drilling as well as unexpected situations. Additionally,
the research by Jeong \textit{et al.} \cite{Jeong2013} and Navabi
\textit{et al.} \cite{Navabi25} introduced advanced sensors with
high cost, which makes the methods impractical to implement broadly.

Our previous work \cite{Zhao2023}\textcolor{blue}{{} }developed an
autonomous robotic drilling system for mice cranial window creation
with only image-based completion level recognition as the perception
on global information and preliminarily proved its feasibility by
eggshell drilling experiment. However, the lack of local contact signals
still results in some limitations in lag in updating recognized penetration
of occlusion areas by drill and a lack of sufficient accuracy, particularly
for pixels with drilling completion levels higher than $0.8$. Jia
\textit{et al.} \cite{Jia2023} presented a 3-phase periodic Bayesian
reinforcement learning method for the eggshell drilling task as a
simulation of cranial window creation and achieved successfully drilling
and automatically acquiring spatial information about the eggshell
terrain via the information of 2D image, force, and audio signal instead
of 3D scanning in advance. This work combined the processing of contact
signals and global awareness along the trajectory and does not require
any pre-operative information input. Regrettably, the work was evaluated
on an egg model rather than on mice, and the Bayesian reinforcement
learning needs a dedicated simulator for training, reducing its practicality.

In summary, the aforementioned studies either focus on local penetration
detection or rely on pre-operative information, with few effectively
combining the two, which limits their real-time adaptability to skull
variations. Some of these works employ specialized sensors or simulations,
affecting their broad practicality. It is also worth noting that except
for \cite{Zhao2023}, the related studies did not consistently report
experimental metrics of drilling speed or success rate, which makes
it difficult for direct comparison. Considering the limitations of
previous studies, we aim to propose an autonomous robotic drilling
system for mice cranial window creation that can perceive both contact
signals and global information without pre-processing, specialized
sensors, or simulators. To achieve this, our system primarily relies
on image processing using a neural network for perception, while force
signals serve as auxiliary inputs. Additionally, we conduct repeat
experiments and evaluate the results with general metrics. The contrast
of the capabilities of the proposed work with existing literature
is shown in Table~\ref{tab:related work}.

\begin{figure}[tbh]
\begin{centering}
\includegraphics[width=0.5\columnwidth]{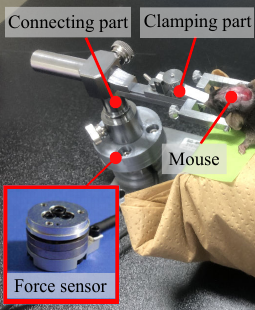}
\par\end{centering}
\caption{\label{fig:clamping platform mice}The clamping mechanism for mouse
skulls including a commercial mouse skull holder and a force sensor.}
\end{figure}

\begin{figure}[tbh]
\begin{centering}
\includegraphics[width=1\columnwidth]{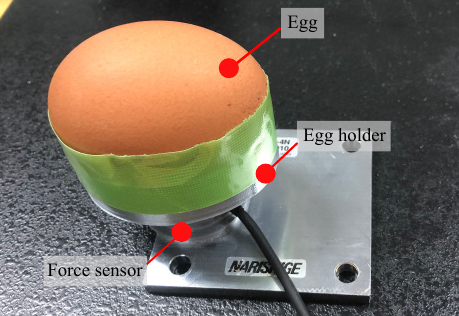}
\par\end{centering}

\caption{\label{fig:clamping platform egg}The clamping mechanism for eggs
including a 3D printed holder and a force sensor.}
\end{figure}

\subsection{Statement of contributions\label{subsec:Statement-of-contributions}}

In this work, we show a trajectory planner and a multimodal recognition
method with image and force information, capable of perceiving global
information and contact signals, enabling fully autonomous operation
in \textit{postmortem} mice, in addition to 7.1 min average drilling
(136\% faster) with increased (95\% vs. 80\%) success rate in autonomous
eggshell drilling. In detail, we propose (1) a trajectory planner
based on constrained splines and execution-time plane fitting that
is adjusted in real-time by image and force feedback, (2) a drilling
completion level recognition module based on deep neural networks
with multi-branch architectures to update the planned trajectory at
execution-time with image and force information. The force information
increases the resolution of recognition 10 times and provides information
on drilled areas under occlusion. In addition, we evaluate (3) the
performance of the drilling completion level recognition integrating
image and force information, (4) the impact of new modules via ablation
study in fully autonomous robotic drilling experiments using eggshells,
and (5) present the first-in-the-World fully autonomous results of
cranial window drilling in \textit{postmortem} mice.

\begin{figure*}[t]
\begin{centering}
\includegraphics[width=1\textwidth]{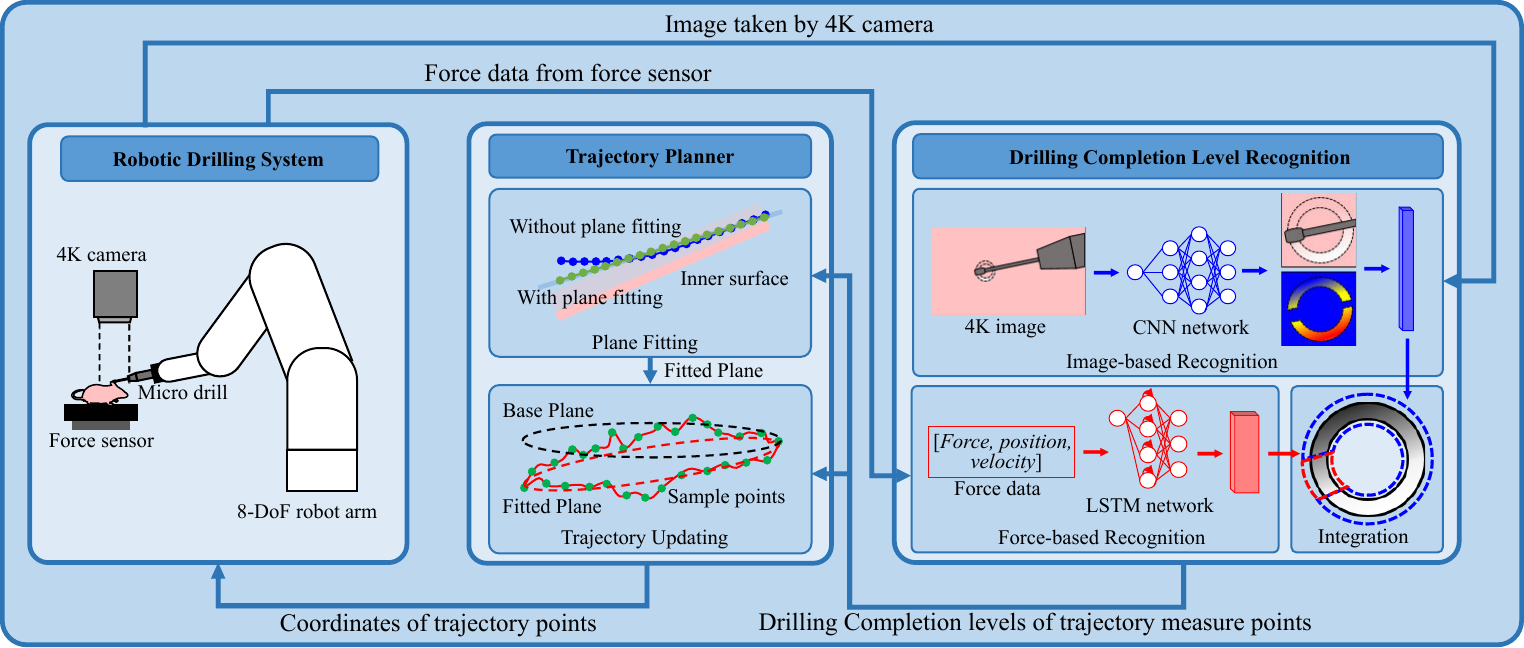}
\par\end{centering}

\caption{\label{fig:overall_system}Block diagram of the proposed autonomous
robotic drilling system.}
\end{figure*}

\section{Problem statement\label{sec:Problem-statement}}

Consider the setup shown in Fig.~\ref{fig:robot_system}, using one
of the robotic arms of our robot platform for scientific exploration
\cite{Marinho2024}. Let $R$ be the 8-degrees-of-freedom serial manipulator
with joint values $\quat q\in\mathbb{R}^{8}$ composed of the robotic
arm (CVR038, Densowave, Japan), linear actuator, and circular rail
actuator. Let $R$ be holding the micro drill (MD1200, Braintree Scientific,
USA) that is used to drill the object (i.e. either the eggshell or
the mouse skull) fixed by our clamping mechanism. The clamping mechanism
contains a force sensor (ThinNANO, BL AUTOTEC, Japan). For mice, the
clamping part is a head holder (SG-4N, NARISHIGE, Japan), shown in
Fig.~\ref{fig:clamping platform mice}. And for eggs it is a custom-designed
3D printed egg holder. The force sensor has a custom-designed 3D printed
casing to fix the force sensor underneath the egg holder, shown in
Fig.~\ref{fig:clamping platform egg}. Images are obtained from above
through a 4K microscope system (STC-HD853HDMI, Omron-Sentech, Japan)
equipped with a distortionless macro lens of $f=75$ mm (VS-LDA75,
VS Technology, Japan), set vertically above the center of the mouse
or egg to observe the drilling procedure. Only for the system of eggshell
drilling, an air compressor (Model OF301-4B, Jun-Air, USA) with a
silicone tube is applied to blow away the shell dust emitted from
drilling the surface of the eggshell.

\subsection{Goal}

Our goal is to autonomously drill the object (eggshell or mouse skull)
along a circular path with respect to the base frame of the robot
system, $\fdrill$ (see Fig.~\ref{fig:robot_system}), without requiring
target-specific data collection, pre-training with accurate simulation,
pre-processing, or surface measurement. In addition to proprioceptive
information (kinematics and encoders), the goal is to use accessible
sensors (camera and force sensor) to gather information for execution-time
feedback, including global information, primarily for monitoring the
drilling process, as well as contact signals to support predictions
in occluded regions. The method can achieve an effect similar to a
dynamically changing visual virtual fixture, without any pre-settings.

\subsection{Overview of the proposed system\label{subsec:Overview-of-the}}

The overview of the proposed system is shown in Fig.~\ref{fig:overall_system}.
In our strategy, the robotic system block interacts with the object
(eggshell or mouse skull) while global information and contact signals
of the drilling procedure are acquired by sensors (4K camera and force
sensor) assembled in the multimodal robotic system. The information
is processed and converted into a generic dynamic parameter, the drilling
completion level $\myvec c$, by the drilling completion level recognition
block (introduced in Section~\ref{sec:Drilling-completion-level}).
$\myvec c$ is an $n$-dimensional vector in which each element $\mathbb{R}\ni c_{i}\triangleq c_{i}\left(t\right)\in\left[0,1\right]$
stands for the completion percentage of drilling at the $i$-th measure
point along the trajectory, where $c_{i}=0$ means not drilled and
$c_{i}=1$ means completely drilled. The drilling completion level-based
trajectory planner (introduced in Section~\ref{sec:Drilling-completion-level-1})
updates the continuous trajectory at execution time. The robot's tooltip
trajectory is calculated by the constrained inverse kinematics algorithm
\cite{Marinho2024}, which outputs joint commands to the robotic system,
closing the loop.

\section{Drilling completion level recognition\label{sec:Drilling-completion-level}}

The drilling completion level recognition block consists of three
parts: namely, image-based completion level recognition (Section~\ref{subsec:Image-based-Completion-Level}),
force-based completion level recognition (Section~\ref{subsec:Force-based-Completion-Level})
and completion level results integration (Section~\ref{subsec:Completion-Level-Result}).
The image-based and force-based completion level recognition parts
can independently recognize the drilling completion levels ($\myvec c_{\text{image}}$
and $\myvec c_{\text{force}}$) for selected measure points on the
trajectory, and the two results are integrated into $\myvec c$ for
improved performance and higher resolution.

\subsection{Image-based completion level recognition\label{subsec:Image-based-Completion-Level}}

\begin{figure*}[tbh]
\begin{centering}
\includegraphics[width=1\textwidth]{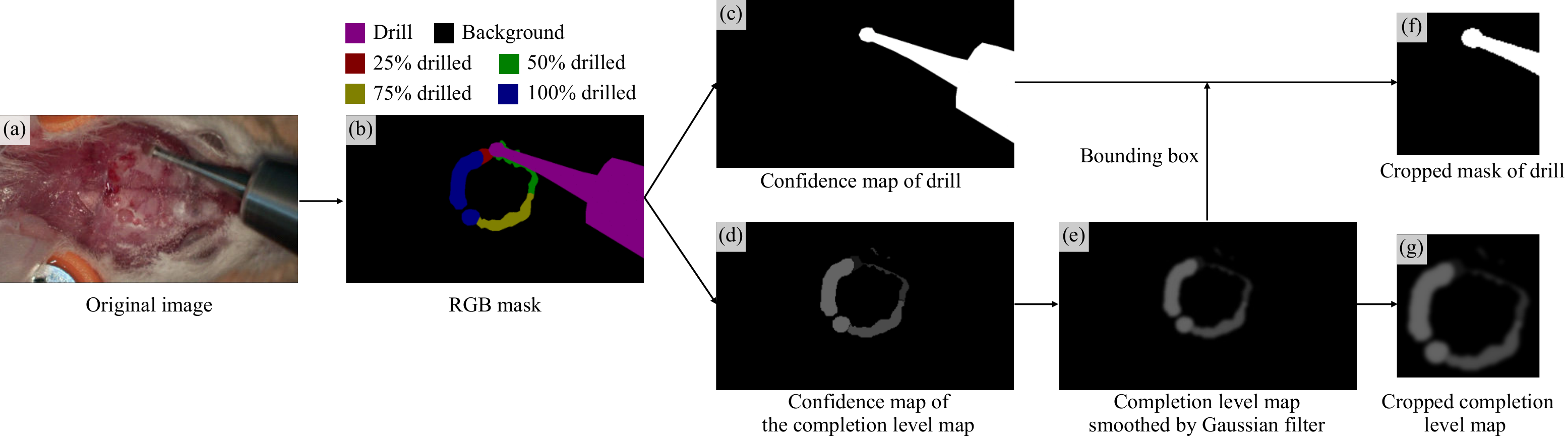}
\par\end{centering}

\caption{\label{fig:dataset}An example of creating the training dataset. (a)
An original image of mouse skull drilling. (b) RGB mask that is labeled
manually in 6 classes: Drill, $0\,\%$ drilled (Background), $25\,\mathrm{\%}$
drilled, $50\,\mathrm{\%}$ drilled, $75\,\mathrm{\%}$ drilled, and
$100\,\mathrm{\%}$ drilled. (c) Confidence map of the drill. (d)
Confidence map of the completion level map. (e) The completion level
map is smoothed by a Gaussian filter. (f) Cropped drill mask within
the drilling region. (g) Cropped completion level map within the drilling
region.}
\end{figure*}
For image-based completion level recognition, we propose a single
network to jointly detect the drilling area and estimate the drilling
completion level, leveraging the synergistic benefits known from co-training
complementary task such as detection and semantic segmentation \cite{Dvornik2017}.
The algorithm and network remain unchanged from the previous study
\cite{Zhao2023}, whose architecture is inspired by DSSD \cite{Fu2017}
and contains two branches for bounding-box detection and completion
level estimation. The main improvement is that in this paper, a mouse
skull dataset is built and applied to retrain the network in preparation
for mouse skull drilling experiments.\textbf{ More training details
(network architecture, hyperparameter tuning, evaluation of joint
training, and training condition) are available in the supplementary
material. }

\subsubsection{Training dataset creation}

The datasets were specifically generated for this task. In our former
study \cite{Zhao2023}, we collected 518 images from videos of teleoperated
robotic eggshell drilling experiments performed by our group on 12
eggs that comprised the eggshell dataset and expanded it to 16,576
images after augmentation methods of random flip, rotate, random crop,
and random change of brightness and contrast. In order to apply the
network to the mouse skull, we collected 587 images from videos of
teleoperated mouse skull drilling experiments performed by our group
on 10 \emph{postmortem} mice for the mouse skull dataset (18,784 after
augmentation). Both datasets were annotated manually and split $80\,\%$-$10\,\%$-$10\,\%$
for training, validation, and testing.

\paragraph{Data annotation}

Generally, ground truth images with full manual annotation are required
for multi-task learning. In this work, we expect our network to output
a bounding box of the drilling area and a completion level map. When
generating the ground truth of completion level maps, manually annotating
each pixel with a grayscale value ranging from 0 to 100 is not feasible
(0 of grayscale refers to the completion level of that pixel $c=0$
while 100 refers to $c=1$). We propose a more treatable annotation
strategy.

To describe the annotation procedure, let us use a mouse skull drilling
image, but the process is the same for eggshell images. As shown in
Fig.~\ref{fig:dataset}-(a), given an image, first we define 6 classes
for a semantic segmentation classifier: drill, $0\,\%$ drilled (background),
$25\,\mathrm{\%}$ drilled, $50\,\%$ drilled, $75\,\%$ drilled,
and $100\,\%$ drilled as shown in Fig.~\ref{fig:dataset}-(b). The
percentage of drilling completion is defined subjectively based on
the video with our experience in pilot studies. Second, at a higher
level in the hierarchy, we separate the image into two confidence
maps. The first one is for pixels of the drill, as shown in Fig.~\ref{fig:dataset}-(c),
with 255 corresponding to the full confidence, so that the drill can
be removed from the mask. The second confidence map corresponds to
the completion level map, as shown in Fig.~\ref{fig:dataset}-(d),
where 0 of grayscale corresponds to $0\,\%$ drilled pixels and 100
corresponds to $100\,\%$ drilled pixels. Lastly, taking advantage
of the expected continuity in drilled regions, a Gaussian filter is
applied to smooth the completion level map channel so that its grayscale
values of pixels are continuous from 0 to 100, as shown in Fig.~\ref{fig:dataset}-(e).

The bounding box is automatically annotated. A 4-dimensional vector
($x_{1},\,y_{1},\,x_{2},\,y_{2}$) is used as the ground truth of
the bounding box detection branch, which is generated from the completion
level map channel by noting the minimum value and maximum value from
width and height direction of those pixels whose grayscale values
are not 0. After that, we crop both the completion level map channel
and drill channel using the coordinate of the bounding box ($x_{1},\,y_{1}$)
and ($x_{2},\,y_{2}$) and then resize them into $128\times128$,
satisfying the requirement of real-time processing. Merging two channels
together we can get a $128\times128\times2$ matrix as the ground
truth of the completion level estimation branch, as shown in Fig.~\ref{fig:dataset}-(f)(g).

\subsubsection{Training result and evaluation\label{subsec:Training-result-image}}

\begin{figure}[tbh]
\begin{centering}
\includegraphics[width=1\columnwidth]{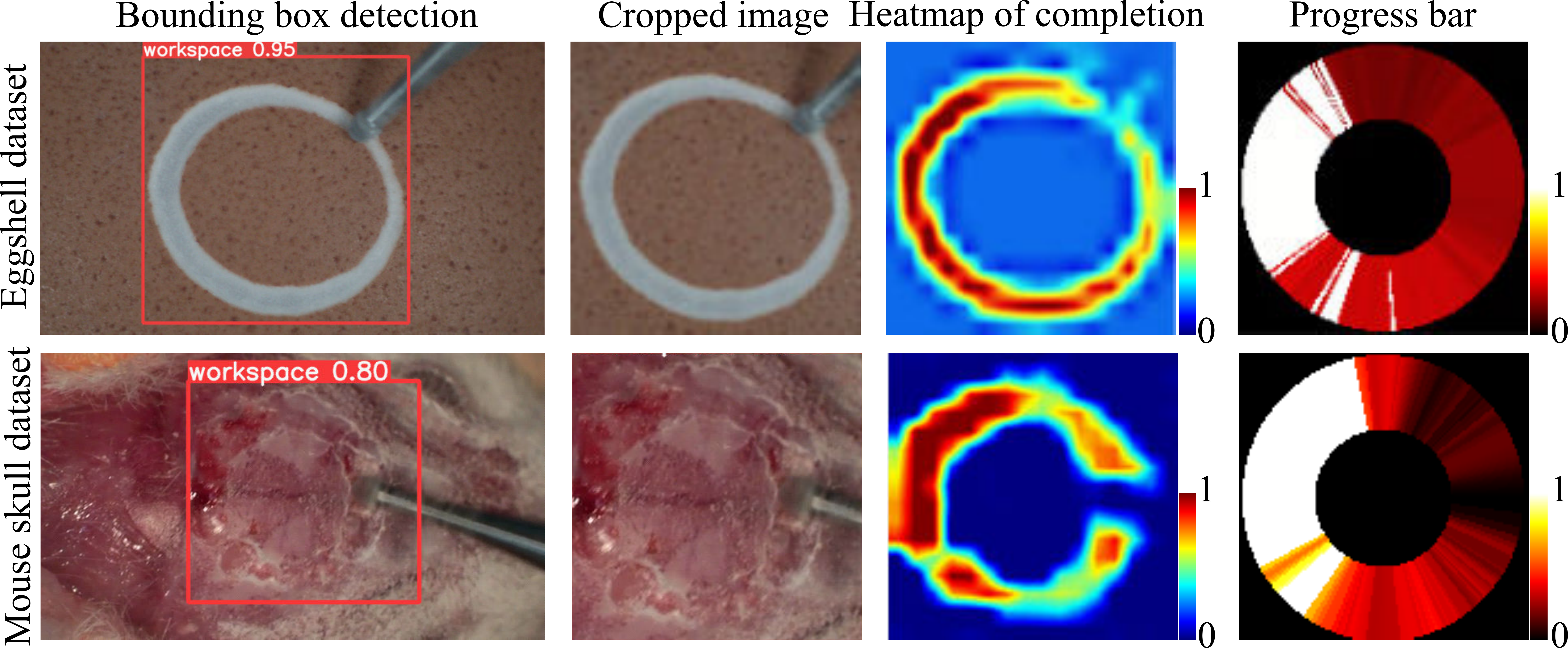}
\par\end{centering}

\caption{\label{fig:result}Output results from the network. The original image
with the bounding box, the cropped image, the heatmap, and the progress
bar of the completion level map are outputs for post-processing and
visualization. Color bars are attached to the heatmap and the progress
bar, which will be omitted in subsequent images.}
\end{figure}

The output results of an eggshell image and a mouse skull image are
shown in Fig.~\ref{fig:result}. The bounding box detection and cropped
image show the output from the network for the detected drilling area
and the cropped result of the area. The heatmap of completion shows
a visualization of the drilling completion levels recognition result
of parts that are drilled and not occluded by the drill, while untouched
or occluded parts are suppressed.

mAP (mean Average Precision) is applied as the evaluation metric for
object detection while MAPE (Mean Absolute Percentage Error) is applied
as the evaluation metric of completion level estimation task. Our
model is able to reach up 78.5 in mAP for detection and $15.05\,\%$
in MAPE for estimation while the frequency is $72$ for the eggshell
dataset. For the mouse skull dataset, the speed is unchanged, the
mAP for detection is 77.6, and the MAPE for estimation is $24.32\,\%$.
Based on the experiment results in our former research \cite{Zhao2023},
reasonable values in this application are $\text{mAP}\,>\,75$, $\text{MAPE}\,<\,25\,\%$
and real-time (FPS$\,>\,60$), so our work is able to achieve a good
trade-off.

We also obtain a progress bar of the drilling procedure by normalizing
the corresponding pixel value. The progress bar is generated by not
only the completion level map in this moment but also those in the
past because we expect the drilling progress of an area to be monotonically
increasing. By setting $m$ discrete points evenly along the circle
on the progress bar, we obtain a $\myvec c_{\text{image}}$ with size
of $m$ and elements $c_{i,\text{image}}\in\left[0,1\right]\subset\mathbb{R}$,
namely the image-based drilling completion level at any point $i$,
where $i=1,\cdots,m\in\mathbb{N}$. The value of $m$ is 32 considering
the resolution of the image.

\subsection{Force-based completion level recognition\label{subsec:Force-based-Completion-Level}}

The force-based completion level recognition is an addition over our
previous research \cite{Zhao2023}. In order to compensate for the
inaccuracy of image processing in estimating results when the drilling
completion level is higher than $0.8$, to address the occlusion of
points by the drill, and to increase spatial resolution, we added
a force-based Long Short-Term Memory Recurrent Neural Network (LSTM-RNN)
\cite{hochreiter1997long}. This approach takes advantage of the higher
spatial accuracy of the force information (w.r.t. image information)
and the significant difference in stiffness between the membrane and
the skull or eggshell. We utilize force data collected during the
drilling procedure to predict the drilling completion level at the
current and subsequent drilling regions.

We posit that recognizing the drilling completion levels in the vicinity
of the drill-surface contact area can be achieved by analyzing the
measurement values of the force sensor and the position and velocity
state of the drill over a short history (e.g., the past few seconds).
To implement this recognition, we employ an LSTM-RNN whose structure
is inspired by \cite{He2019}. \textbf{More training details (network
architecture and training conditions) are available in the supplementary
material.}

\subsubsection{Training dataset}

The proposed LSTM-RNN takes force, positions, and velocities of the
drill in past $l$ seconds as the input, and outputs the drilling
completion levels from time $t$ to $t+k$ in the future as the drill
moves a complete circular trajectory on the $x\text{--}y$ plane.
Note that $t$ denotes the current moment, and $k$ is the predictable
number of future time. Assuming that the sampling frequency of the
force sensor is $f$, the sample size of the input data is $L=fl$,
and the sample size of output data is $K=fk$. Considering our drilling
speed of 16 seconds per turn, we chose $l=16$ and $k=4$ considering
the size of the occlusion area. The sampling frequency of the force
sensor was set as $f=20$ Hz. Thus the sample sizes of input data
and output data are $L=320$ and $K=80$, respectively.

To train the model, we conducted 10 autonomous (image-only) drilling
trials each on eggshells and mice, recording force $F_{x},F_{y},F_{z}$,
position $x,y,z$, descent speed $v_{z}$, and drilling completion
level $c$. The ground truth for $c$ was obtained through the manual
annotation (see Fig.~\ref{fig:dataset}) and upsampled to 320 points
to match the sensor data resolution.

Each trial produced an $S\times8$ matrix ($S$ being the total samples),
which was cut into $S-L-K$ sets of size $L\times7$ (input) and $K\times1$
(output) for training. The eggshell dataset contained 84,000 samples
(80,000 sets after cutting), and the mouse skull dataset had 87,600
samples (83,600 sets after cutting). Both datasets were split $80\,\%$-$10\,\%$-$10\,\%$
for training, validation, and testing.

\subsubsection{Training result and evaluation\label{subsec:Training-result-force}}

The LSTM-RNN is expected to output the prediction result of drilling
completion levels from time $t$ to $t+k$, so we can expect that
predicted drilling completion level at time $t+\Delta t$ ($0\leqslant\Delta t\leqslant k$)
should be an implicit variable in $\Delta t$. By comparing the prediction
result at time $t+\Delta t$ with the ground truth, the prediction
accuracy of the network for $\Delta t$ time forward can be calculated,
denoted as $acc\left(\Delta t\right)$. In our case, we allow a difference
between predicted and true value within $0.05$.

\begin{figure}[tbh]
\begin{centering}
\includegraphics[width=1\columnwidth]{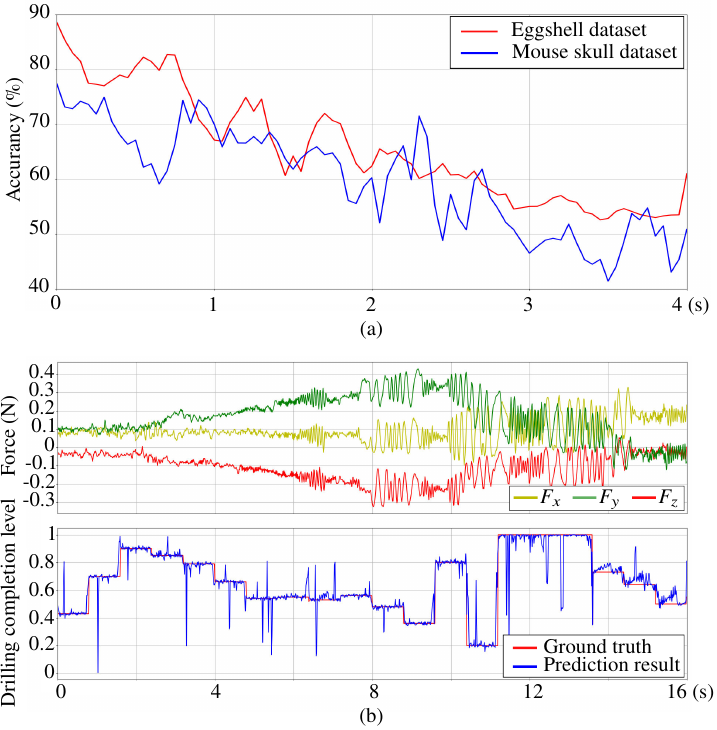}
\par\end{centering}

\caption{\label{fig:force_accu}Training results of the LSTM-RNN. (a) The image
of the change in accuracies of completion level recognition $acc\left(\Delta t\right)$
with $\Delta t$ for eggshell dataset (red line) and mouse skull dataset
(blue line) based on data from time $t-l$ to $t$; (b) Forces landscape
and the prediction result at time $t$ within a drilling turn.}
\end{figure}

The curves of changes in accuracy $acc\left(\Delta t\right)$ for
the eggshell dataset and the mouse skull dataset are illustrated as
the red and blue lines in Fig.~\ref{fig:force_accu}-(a). When $\Delta t=0$,
the accuracies of both curves reach their maximum value, $88.60\,\%$
for the eggshell dataset and $77.43\,\%$ for the mouse skull dataset.
As $\Delta t$ increases, both curves show a decreasing trend. For
the eggshell dataset, the lowest accuracy $52.61\,\%$ occurs when
$\Delta t=3.45$, while for the mouse skull dataset, the lowest accuracy
$41.21\,\%$ is at $\Delta t=3.50$. The result is in line with our
predictions because larger $\Delta t$ is farther into the future,
which is more difficult for the neural network to predict. By linear
fitting of the accuracy using the least squares method, we can obtain
an expression for the accuracy
\begin{equation}
acc\left(\Delta t\right)=a\cdot\Delta t+b\:(\%).\label{eq:accu}
\end{equation}
For the eggshell dataset, the values of parameters $a$ and $b$ are
$a=-7.94$, $b=81.43$. For the mouse skull dataset, the parameter
values are $a=-7.07$ and $b=74.01$. The variation in accuracy influences
the integration strategy, which will be discussed in Section~\ref{subsec:Completion-Level-Result}.
Additionally, an example of the forces landscape and the prediction
completion level result when $\Delta t=0$ is shown in Fig.~\ref{fig:force_accu}-(b).

The output of the LSTM-RNN $\myvec c_{\text{force}}$ is a vector
with size $K=80$ and each element value $c_{i,\text{force}}\in\left[0,1\right]\subset\mathbb{R}$
refers to the drilling completion level at at a future moment $t+\nicefrac{i}{f}$
in terms of temporal span.

\subsection{Completion level results integration\label{subsec:Completion-Level-Result}}

In Section~\ref{subsec:Image-based-Completion-Level} and Section~\ref{subsec:Force-based-Completion-Level},
the drilling completion levels were independently recognized based
on image and force information, albeit at different spatial and temporal
scales. The advantage of image-based recognition is that the image
taken by the upper camera provides single-shot drilling completion
levels of all points along the trajectory, while the disadvantages
are relatively low spatial resolution and occlusion issues. On the
other hand, force-based recognition can predict the instantaneous
changes in drilling completion levels of the contact area at the current
moment with high accuracy and resolution without occlusion issues
but has little global prediction power besides a close future. In
order to make good use of the advantages of both recognition methods,
in this work we propose an integration strategy.

\subsubsection{Synchronization}

Our system uses ROS Noetic Ninjemys for data acquisition, distribution,
and communication, including advanced message synchronization to align
timestamps between 4K camera images and force sensor data, addressing
temporal inconsistencies in sensor fusion \cite{Li2023}. However,
harmonizing the differing lengths and sample rates from sensors remains
necessary.

The image-based completion level output, $\myvec c_{\text{image}}$,
has $m=32$ measure points (described in Section~\ref{subsec:Image-based-Completion-Level}),
while the force-based output, $\myvec c_{\text{force}}$, has $K=80$
points (described in Section~\ref{subsec:Force-based-Completion-Level}).
By padding $\myvec c_{\text{force}}$ with zeros and upsampling $\myvec c_{\text{image}}$,
we expand the size of both vectors to $n=320$. This ensures that
both vectors have the same size, with each element corresponding to
the drilling completion level at the same trajectory point based on
image and force data. \textbf{More details on synchronization are
available in the supplementary material.}

\subsubsection{Data fusion}

As mentioned in Section~\ref{subsec:Overview-of-the}, our ultimate
goal is to obtain the drilling completion level $\myvec c$ with size
of $n$, generated by the fusion of image and force. The completion
level $\myvec c$ can be calculated as
\begin{equation}
\myvec c=\myvec w_{1}\odot\myvec c_{\text{image}}+\myvec w_{2}\odot\myvec c{}_{\text{force}},\label{c=00003Dwc+wc}
\end{equation}
where $\odot$ denotes the element-wise product of two vectors, and
$\quat w_{1}$ and $\quat w_{2}$ are the weight vectors with size
of $n$ whose elements are the weight of the corresponding element
in $\myvec c_{\text{image}}$ and $\myvec c{}_{\text{force}}$.

The force-based accuracy weights, $\quat w_{2}$, are derived using
the accuracy function $acc_{i}=a\cdot\Delta t_{i}+b\:(\%)$, where
$\Delta t_{i}=\nicefrac{i}{f}$, $i=1,\cdots,K\in\mathbb{N}$, and
$f$ is the force sensor frequency. The parameters $a$ and $b$ are
set as $-7.94$, $81.43$ for eggshells and $-7.07$, $74.01$ for
mouse skulls (discussed in Section~\ref{subsec:Force-based-Completion-Level}).
Similar to $\myvec c_{\text{force}}$, we also fill $\quat w_{2}$
with extra elements of zeros, expanding its size to $n=320$. Whenever
there is occlusion, the image-related results are less reliable, therefore
we set $\quat w_{1}=1-\quat w_{2}$. Substituting the results of $\quat w_{1}$
and $\quat w_{2}$ into \eqref{c=00003Dwc+wc}, the integration result
$\myvec c$ can be calculated.

Compared with our previous research \cite{Zhao2023}, with the integration
of force information, the accuracy of the completion level recognition
improves close to the current drilling area and the spatial resolution
of the completion level recognition increases 10 fold. The recognition
result of the multimodal integration is used in \eqref{eq:vp}.

\section{Drilling completion level-based trajectory planner\label{sec:Drilling-completion-level-1}}

In the last section, the drilling completion level recognition block
outputs the integration result of the completion level $\myvec c$
that is a discrete set of $n$ values ($n=32$ for \cite{Zhao2023}
using only image and $n=320$ for the proposed image and force module).
Nonetheless, the robot must be given a continuous trajectory for smooth
performance using our inverse kinematics calculation \cite{Marinho2024}.
In this section, we present a trajectory planner that generates a
continuous trajectory for autonomous drilling, where each point on
the trajectory has a fixed coordinate value in the $x\text{--}y$
plane while the $z-$axis coordinate is updated based on the drilling
completion level.

In formal terms, let the desired drill path be discretized into $n$
equal intervals resulting in the points $\quat p_{i}\left(c_{i}\right)\triangleq\begin{bmatrix}p_{x,i} & p_{y,i} & p_{z,i}\left(c_{i}\right)\end{bmatrix}^{\text{T}}\in\mathbb{R}^{3}$,
$i=1,\cdots,n\in\mathbb{N}$, defined with respect to the base frame
of the robotic manipulator $\fdrill$ (see Fig.~\ref{fig:robot_system}).
$c_{i}$ is the drilling completion level of each point $i$ recognized
from the image and force, detailed in Section~\ref{sec:Drilling-completion-level}.
The goal of the trajectory planner is to obtain a continuous trajectory
that is circular \emph{when projected} in the $x\text{--}y$ plane
and causes all $n$ drill points to be complete in finite time by
updating the $z-$axis position, $p_{z,i}$, using the drilling completion
level $c_{i}$. We hypothesize that, by using a proper interpolation
methodology, if all $n$ points (the ones we can effectively measure
online) are sufficiently drilled, then the entire trajectory (including
the points that we cannot directly measure) will be sufficiently drilled.

The proposed trajectory planner is divided into three steps. First,
imbued with each drilling completion level (obtained in Section~\ref{sec:Drilling-completion-level}),
$c_{i}$, we calculate each $p_{z,i}\left(c_{i}\right)$ as discussed
in Section~\ref{subsec:Z-coordinate-Value-Calculation}. Second,
a plane is fitted based on those points $\quat p_{i}\left(c_{i}\right)$
whose drilling completion level $c_{i}\neq0$, obtaining an offset
$o_{z,i}$ between each point and its projection on the plane to update
$p_{z,i}\left(c_{i}\right)$ on the trajectory as discussed in Section~\ref{subsec:Plane-Fitting}.
Lastly, with the $z-$axis coordinate for each of the $n$ measure
points defined in the previous step and given the desired trajectory
topology, we obtain the continuous drill path using a constrained
spline interpolator as discussed in Section~\ref{subsec:Constrained-Cubic-Spline}.

\subsection{$z-$axis trajectory calculation: multimodal velocity damper\label{subsec:Z-coordinate-Value-Calculation}}

The core of our automatic drilling proposal is to reduce the downward
velocity $v_{z,i}\triangleq v_{z,i}\left(c_{i}\right)\in\mathbb{R}$
of a given trajectory point $i$ proportionally to its drilling completion
level $0\leqslant c_{i}\leqslant1$, obtained via multimodal information
as shown in Section~\ref{sec:Drilling-completion-level}. The velocity
is modulated as follows 
\begin{align}
v_{z,i}\left(c_{i}\right)= & \left(1-c_{i}\right)v_{z}.\label{eq:vp}
\end{align}
This means that when a trajectory point $i$ has drilling completion
level $c_{i}=0$, it is untouched, hence $v_{z,i}\left(0\right)=v_{z}$
and the robot moves at nominal velocity $v_{z}$ downward, where $v_{z}$
is a design parameter with a negative value (e.g., -1 mm/s), since
the upward direction is defined as the positive direction in the base
frame of the robot system $\fdrill$. When the trajectory point has
$c_{i}=1$, that point has been fully drilled and the downward velocity
becomes $v_{z,i}\left(1\right)=0$, meaning that the robot will not
drill further. Intermediate levels of completion allow the system
to behave smoothly with respect to the drilling completion level.

Although effective, a direct implementation of \eqref{eq:vp} has
been shown in our previous work \cite{Zhao2023} to be inefficient
when the target circular drill surface is not aligned with the robot's
$x\text{\textendash}y$ plane.

In this work, we augment \eqref{eq:vp} with an online estimation
of the target $x\text{\textendash}y$ drill plane using $c_{i}$,
described in Section~\ref{subsec:Plane-Fitting}. Considering the
simple integration of \eqref{eq:vp} with a sampling time $T$, the
plane-fitted position-based algorithm becomes
\begin{equation}
p_{z,i}\left(t+T\right)=p_{z,i}\left(t\right)+v_{z,i}\left(c_{i}\right)T+o_{z,i}(t),\label{eq:zp}
\end{equation}
where $o_{z,i}(t)$ is the plane-fitted offset, detailed in Section~\ref{subsec:Plane-Fitting}.
This way, the drilling traverses through empty space much faster by
estimating the surface plane of the target and adapting accordingly.

\subsection{Continuous trajectory generation: constrained cubic spline interpolation\label{subsec:Constrained-Cubic-Spline}}

\begin{figure}[tbh]
\begin{centering}
\includegraphics[width=1\columnwidth]{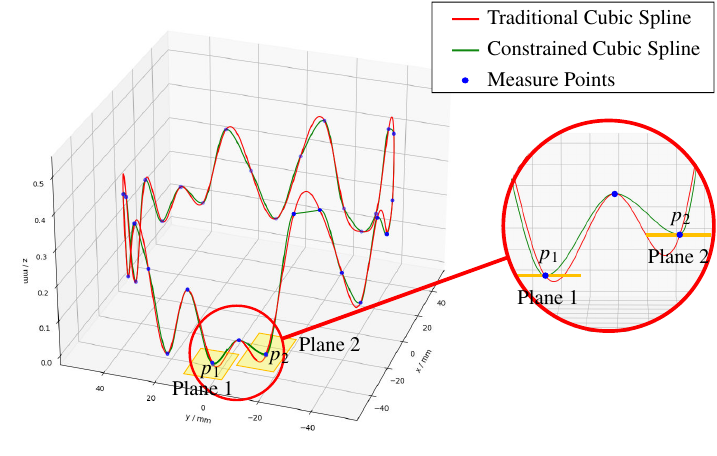}
\par\end{centering}

\caption{\label{fig:curve}Generated curve based on a certain amount of measure
points along the circular path. The blue points are the measure points,
the red line is the result of traditional spline interpolation, and
the green line is the result of constrained spline interpolation.
We can see the red line overshoots while the green does not.}
\end{figure}

Since the $z-$axis coordinate for each of the $n$ measure discrete
points $p_{z,i}\left(c_{i}\right)$ has been updated in the previous
section, the purpose of this section is to generate a continuous drilling
path based on these points to match the high precision encoders used
in the robotic system. Cubic spline interpolation has been widely
applied for generating continuous paths from discrete points but it
is, by itself, unsuitable for our application as it might overshoot
between two trajectory measure points and cause rupture to the membrane
(see Fig.~\ref{fig:curve}). Instead, we propose the use of constrained
spline interpolation by eliminating the requirement for equal second
order derivatives at every point and replacing it with specified first
order derivatives \cite{kruger2003constrained}. As an improvement
over our previous work \cite{Zhao2023}, we focus on the relevant
coordinates given that our trajectory is cylindrical.

Let the trajectory points be $\begin{bmatrix}p_{x,i} & p_{y,i} & p_{z,i}\end{bmatrix}^{\text{T}}$,
$i=1,\cdots,n\in\mathbb{N}$. We express them under a cylindrical
coordinate system $\begin{bmatrix}p_{\rho,i} & p_{\varphi,i} & p_{z,i}\end{bmatrix}^{\text{T}}$.
The $z-$axis coordinate $p_{z,i}$ is the same for both expressions
while
\begin{equation}
\begin{array}{c}
p_{\rho,i}\triangleq\sqrt{p_{x,i}^{2}+p_{y,i}^{2}}=\frac{d}{2}\\
p_{\varphi,i}={\rm atan2}\left(p_{y,i},p_{x,i}\right)
\end{array},\label{eq:cylindrical}
\end{equation}
where $d$ is the diameter of the cylinder, which is a constant value
of 8 mm in our setup and the function ${\rm atan2}\left(y,x\right)$
calculates the polar angle of the point $\left(x,y\right)\neq\left(0,0\right)$
taking values in $(-\pi,\pi]$. As a result, interpolation happens
in $\mathbb{R}^{2}$, namely the points $\quat{\rho}_{i}=\left[\begin{array}{cc}
p_{\varphi,i} & p_{z,i}\end{array}\right]^{\text{T}}$ with $p_{\varphi,i}\in[-\pi,\pi)\subset\mathbb{R}$ and $p_{z,i}\in\mathbb{R}$.

Let $\varphi$ be the independent variable to generate a curve $\quat s\left(\varphi\right)=\left[\begin{array}{cc}
\varphi & s_{z}\left(\varphi\right)\end{array}\right]^{\text{T}}$ through $[-\pi,\pi)$. We have then a set of piecewise curves $\quat s_{i}\left(\varphi\right)=\left[\begin{array}{cc}
\varphi & s_{z,i}\left(\varphi\right)\end{array}\right]{}^{\text{T}}$, each $i$-th curve describing a curve between two contiguous points
$\quat{\quat{\rho}}_{i}$ and $\quat{\quat{\rho}}_{i+1}$. By the
definition of a cubic spline, each curve is a third-degree polynomial
given by 
\begin{equation}
s_{z,i}\left(\varphi\right)\triangleq a_{z,i}+b_{z,i}\varphi+c_{z,i}\varphi^{2}+d_{z,i}\varphi^{3},\label{eq:cons1}
\end{equation}
whose first-order derivative of \eqref{eq:cons1} is
\begin{equation}
\dot{s}_{z,i}\left(\varphi\right)\triangleq b_{z,i}+2c_{z,i}\varphi+3d_{z,i}\varphi^{2}.\label{eq:cons2}
\end{equation}

In order to calculate the parameters of the curve, we need three continuity
conditions. First, given that the domain for each $\quat s_{i}\left(\varphi\right)$
is $[p_{\varphi,i},p_{\varphi,i+1}]$, each piecewise curve must pass
through its ``left'' endpoint $\quat p_{i}$ and ``right'' endpoint
$\quat p_{i+1}=\left[\begin{array}{cc}
p_{\varphi,i+1} & p_{z,i+1}\end{array}\right]^{\text{T}}$, that is,
\begin{equation}
s_{z,i}\left(p_{\varphi,i}\right)\triangleq p_{z,i},\label{eq:con3}
\end{equation}
\begin{equation}
s_{z,i}\left(p_{\varphi,i+1}\right)\triangleq p_{z,i+1}.\label{eq:cons4}
\end{equation}

Second, the first-order derivative of the whole curve must be continuous,
which means the whole curve is continuously differentiable at each
$\quat p_{i}$. Its first-order derivative, $\dot{p}_{z,i}$, is a
free parameter. In traditional cubic spline interpolation, the third
condition is to define the second-order derivative for each measure
point as $\ddot{p}_{z,i}$. As a result, the generated trajectory
is smooth enough but overshoot might occur.

For constrained cubic spline interpolation, instead, we aim to solve
the overshoot problem by sacrificing smoothness. We constrain the
first-order derivative to prevent overshoot between two contiguous
points as 
\begin{equation}
\dot{s}_{z,i}\left(p_{\varphi,i}\right)\triangleq\dot{p}_{z,i},\label{eq:cons5}
\end{equation}
\begin{equation}
\dot{s}_{z,i}\left(p_{\varphi,i+1}\right)\triangleq\dot{p}_{z,i+1},\label{eq:cons6}
\end{equation}
where the intention is to ensure that the first order derivative at
a point will be between the slope of the two adjacent lines joining
that point, and should approach zero if the slope of either line approaches
zero. Furthermore, the first order derivative of a point should be
0 if the signs of the slope of the two lines are different from each
other to define this point as a local minimum to prevent overshoot.
To define a first order derivative that satisfies the conditions,
we denote the product of the differences between the $z$-coordinates
of a point $p_{z,i}$ and its left point $p_{z,i-1}$ and right points
$p_{z,i+1}$ as
\begin{equation}
t_{i}\triangleq\left(p_{z,i+1}-p_{z,i}\right)\left(p_{z,i}-p_{z,i-1}\right),\label{eq:ti}
\end{equation}
so the first order derivative at the point can be calculated as
\begin{equation}
\begin{array}{ccc}
\dot{p}_{z,i} & \triangleq & \left\{ \begin{array}{cc}
\frac{2}{\frac{p_{\varphi,i+1}-p_{\varphi,i}}{p_{z,i+1}-p_{z,i}}+\frac{p_{\varphi,i}-p_{\varphi,i-1}}{p_{z,i}-p_{z,i-1}}} & \text{if \ensuremath{t_{i}>0}}\\
0 & \text{otherwise}
\end{array}\right.\end{array}.\label{eq:szi}
\end{equation}
It is important to note that \eqref{eq:ti} and \eqref{eq:szi} are
only valid when $2\leqslant i\leqslant n-1$. Considering the projection
of the curve on the $x\text{--}y$ plane is a circle, the first point
and the end point of the trajectory are adjacent. We can define $t_{1}\triangleq\left(p_{z,2}-p_{z,1}\right)\left(p_{z,1}-p_{z,n}\right)$
and $t_{n}\triangleq\left(p_{z,1}-p_{z,n}\right)\left(p_{z,n}-p_{z,n-1}\right)$.
So when $t_{1}>0$ or $t_{n}>0$, \eqref{eq:szi} can be adjusted
to obtain
\begin{equation}
\dot{s}_{z}\left(p_{\varphi,1}\right)=\dot{p}_{z,1}\triangleq\frac{2}{\frac{p_{\varphi,2}-p_{\varphi,1}}{p_{z,2}-p_{z,1}}+\frac{p_{\varphi,1}-(p_{\varphi,n}-2\pi)}{p_{z,1}-p_{z,n}}},\label{eq:sz1}
\end{equation}
\begin{equation}
\dot{s}_{z}\left(p_{\varphi,n}\right)=\dot{p}_{z,n}\triangleq\frac{2}{\frac{(p_{\varphi,1}+2\pi)-p_{\varphi,n}}{p_{1}-p_{z,n}}+\frac{p_{\varphi,n}-p_{\varphi,n-1}}{p_{z,n}-p_{z,n-1}}}.\label{eq:szn}
\end{equation}

We can calculate all coefficients by combining \eqref{eq:cons1},
\eqref{eq:cons2}, \eqref{eq:con3}, \eqref{eq:cons4}, \eqref{eq:cons5},
and \eqref{eq:cons6} into
\begin{equation}
\left[\begin{array}{c}
a_{z,i}\\
b_{z,i}\\
c_{z,i}\\
d_{z,i}
\end{array}\right]=\left[\begin{array}{cccc}
1 & p_{\varphi,i} & p_{\varphi,i}^{2} & p_{\varphi,i}^{3}\\
1 & p_{\varphi,i+1} & p_{\varphi,i+1}^{2} & p_{\varphi,i+1}^{3}\\
0 & 1 & 2p_{\varphi,i} & 3p_{\varphi,i}^{2}\\
0 & 1 & 2p_{\varphi,i+1} & 3p_{\varphi,i+1}^{2}
\end{array}\right]^{-1}\left[\begin{array}{c}
p_{z,i}\\
p_{z,i+1}\\
\dot{p}_{z,i}\\
\dot{p}_{z,i+1}
\end{array}\right].\label{eq:abcd}
\end{equation}

By substituting \eqref{eq:szi}, \eqref{eq:sz1}, and \eqref{eq:szn}
into the right side of \eqref{eq:abcd}, we can get the value of coefficients
expression $a_{z,i}$, $b_{z,i}$, $c_{z,i}$, and $d_{z,i}$. We
can obtain the expression of any piecewise third-degree polynomial
curves $\quat s_{i}\left(\varphi\right)$ and, consequently, know
the entire curve $\quat s\left(\varphi\right)$. The cylindrical coordinates
are mapped back into the Cartesian space $\fdrill$ by the reverse
coordinate transformations of \eqref{eq:cylindrical} before going
into the manipulator's inverse kinematics calculation \cite{Marinho2024}.
Fig.~\ref{fig:curve} shows a qualitative comparison between a traditional
cubic spline interpolation and its constrained version. This is to
show that the green curve is enveloped by the tangent planes but red
the curve is not, hence illustrating the problem of overshoot.

\subsection{\label{subsec:Plane-Fitting}Plane fitting}

As mentioned in the previous work \cite{Zhao2023}, the problem of
the long drilling time and potential failure cases were partially
caused by a tilted initial pose. In this work, we address this issue
with an execution-time plane fitting, which is to find the least squares
plane of the measure points whose drilling completion levels are not
0 (given that they are the only points about which we have enough
information) to accelerate the drilling procedure.

After the plane fitting, the original points on the trajectory should
be updated to the fitted plane. At time $t$, consider a certain point
on the trajectory $\quat p_{i}\left(t\right)\triangleq\begin{bmatrix}p_{x,i}\left(t\right) & p_{y,i}\left(t\right) & p_{z,i}\left(t\right)\end{bmatrix}^{\text{T}}$,
the $z$ coordinate of the corresponding point on the fitted plane
is denoted as $p_{z,i}^{\prime}\left(t\right)$ and the offset $o_{z,i}(t)$
in \eqref{eq:zp} can be calculated by
\[
o_{z,i}(t)=p_{z,i}^{\prime}\left(t\right)-p_{z,i}\left(t\right).
\]

We tested the trajectory planner with plane fitting on part of the
trajectory points in simulation and found that the trajectory adapted
to the surface faster by 43\% with plane fitting, which is in line
with our expectation. \textbf{More details on the mathematical description
and simulation testing of the plane fitting are available in the supplementary
material.}

\section{Experiments}

\begin{table*}[tbh]
\centering{}\caption{\label{tab:comparison-state-of-art-egg}The comparison results of
the proposed method with other state-of-the-art deep learning image
processing models.}
\begin{tabular}{c|cccc|cccc|c}
\multirow{2}{*}{Network} & \multicolumn{4}{c|}{MAPE ($\%$) on egg} & \multicolumn{4}{c|}{MAPE ($\%$) on mice} & \multirow{2}{*}{FPS}\tabularnewline
\cline{2-9}
 & Original & Low blur & Medium blur & High blur & Original & Low blur & Medium blur & High blur & \tabularnewline
\hline 
CABiNet \cite{Kumaar2021} & 14.25 & 18.95 & 23.28 & 31.87 & 25.52 & 28.03 & 28.80 & 51.14 & 77\tabularnewline
\hline 
BiSeNet (Res18) \cite{Yu2021} & 15.20 & 19.20 & 23.21 & 31.89 & 25.61 & 27.78 & 28.85 & 50.35 & 66\tabularnewline
\hline 
PP-LiteSeg-T2 \cite{Peng2022} & 14.42 & 19.00 & 22.76 & 30.81 & 24.01 & 27.53 & 30.20 & 49.35 & 96\tabularnewline
PP-LiteSeg-B2 \cite{Peng2022} & 13.97 & 18.85 & 22.35 & 30.66 & 23.87 & 27.23 & 29.65 & 47.92 & 67\tabularnewline
\hline 
SFNet (DF2) \cite{Li2020} & 12.95 & 18.23 & 22.01 & 30.45 & 22.96 & 27.35 & 30.05 & 46.43 & 88\tabularnewline
\hline 
DDRNet-23-S \cite{Pan2023} & 12.78 & 17.92 & 21.11 & 30.29 & 22.62 & 27.01 & 29.15 & 45.01 & \textbf{108}\tabularnewline
\hline 
PIDNet-S-Simple \cite{Xu_2023_CVPR} & 12.38 & 17.78 & 20.58 & 30.27 & 21.54 & 26.98 & 28.31 & 45.25 & 101\tabularnewline
PIDNet-S \cite{Xu_2023_CVPR} & 12.25 & 17.65 & 20.43 & 30.12 & 21.32 & 26.64 & 27.70 & 44.93 & 93\tabularnewline
\hline 
Image only {[}ours{]} & 14.97 & 19.23 & 22.95 & 31.03 & 23.92 & 27.70 & 29.97 & 51.05 & 73\tabularnewline
Image + Force {[}ours{]} & \textbf{9.47} & \textbf{13.31} & \textbf{14.47} & \textbf{27.89} & \textbf{20.38} & \textbf{24.45} & \textbf{25.36} & \textbf{40.94} & 68\tabularnewline
\end{tabular}
\end{table*}

Comparing with our previous system that recognized the drilling completion
level solely based on image \cite{Zhao2023}, in this study, the force-based
completion level recognition and the plane fitting were introduced.
To evaluate the drilling completion level recognition integrated with
force signals, we compare it with the state-of-the-art deep learning
image processing methods. Then, to verify the impact of the two new
modules on improving the system performance, we conducted ablation
studies on eggshell drilling. Lastly, we use the whole system to conduct
an experiment on \emph{postmortem} mouse skull drilling to validate
its effectiveness. \textbf{A video of the experiments is shown in
the supplementary material and available at \url{https://youtu.be/e1ZEtLM033Y}.}

\subsection{Hardware setup \& preparation}

The hardware of the system is set up as described in Section~\ref{sec:Problem-statement}
and shown in Fig.~\ref{fig:robot_system}. As for the image-based
and force-based completion level recognition models, the eggshell
drilling experiment uses the pre-trained model on the eggshell dataset,
while the mouse skull drilling experiment uses the pre-trained model
on the mouse skull dataset. Before the experiment starts, the micro
drill is teleoperated by an operator to adjust the starting point
so that it is within the field-of-view of the camera throughout the
entire circular path.

\subsection{Software implementation}

The experiments use a software implementation on a Ubuntu 20.04 x64
system. The robotic arm is controlled as described in \cite{Marinho2024}.
ROS Noetic Ninjemys is used for the interprocess communication and
CoppeliaSim (Coppelia Robotics, Switzerland) for the simulations.
Communication with the robot is enabled by the SmartArmStack\footnote{https://github.com/SmartArmStack}.
The dual quaternion algebra and robot kinematics are implemented using
DQ Robotics \cite{adorno2021dqrobotics} with Python3. The force sensor
is connected to a Raspberry Pi 4 Model B and the force data is read
at 128 samples per second (SPS) using Python's SMBus module and an
I2C analog input board. \textbf{More details about the software implementation
are available in the supplementary material.} 

\subsection{Parameters selection\label{subsec:Preparation}}

The parameters used in the experiments, namely the initial speed $v_{z}$
of the drill in \eqref{eq:vp}, the diameter of the circular drilling
path $d$, and the frequency $f$ of the transmission of the calculated
$z-$positions to the robot are selected to be $v_{z}\triangleq-6\times10^{-6}$
m/s based on a comparative experiment\textbf{ (detailed in the supplementary
materials)}, $d\triangleq8$ mm based on the demand for follow-up
experiments and the size of mice, and $f\triangleq30$ Hz based on
the requirement of high precision encoders used in the robotic system.
The drilling will stop autonomously when the system judged the drilling
completion criterion is met. This happens when $80\,\%$ of the points
on the drilling path achieve at least $0.85$ of completion level,
consistent with the literature \cite{Zhao2023}.

\subsection{Experiment 1: evaluation of the drilling completion level recognition}

Our proposed drilling completion level recognition method is based
on the image-based element and the force-based element, evaluated
in Section~\ref{subsec:Training-result-image} and Section~\ref{subsec:Training-result-force},
respectively. Therefore, in this experiment, we evaluate the complete
method and compare it to other state-of-the-art real-time semantic
segmentation models. Note that considering that the target tasks are
different, the output of other semantic segmentation models are adjusted
to be two channels: the completion level map channel and drill channel,
the same as those of our method. Moreover, our proposed image-based
completion level recognition can perform bounding box detection and
completion level estimation simultaneously but other models can only
estimate the drilling completion level. For a fair comparison, we
crop the images manually for the compared works to focus on the drilling
area and resize it into $128\times128$ (the same as the output size
of our completion level estimation branch) to serve as the input for
the evaluation of other models, while the original whole image is
applied directly for our model.

\subsubsection{Evaluation dataset}

Three autonomous drilling trials on egg and one on mouse are conducted
for evaluation, with 4K images obtained from the camera and force
signals obtained from the force sensor. These data are for evaluation
only and are not included in the training dataset. Additionally, in
order to evaluate the robustness of the models, we apply Gaussian
filters with kernel sizes $k=3$, $k=5$, and $k=7$ to the images,
adding low, medium, and higher blur, respectively.

\subsubsection{Metrics}

Same as Section~\ref{subsec:Training-result-image}, MAPE is the
accuracy metric and FPS is recorded to evaluate the computation speed. 

\subsubsection{Results and discussion}

The comparison results are summarized in Table~\ref{tab:comparison-state-of-art-egg}.
Noting that we require the drilling completion level recognition to
provide a real-time feedback, so only the result of models with FPS$\,>\,60$
are shown. We can conclude from the result that with the integration
of force signals, our method of drilling completion level recognition
has the best performance and robustness in MAPE while satisfying the
requirement of real-time.

\subsection{Experiment 2: ablation study of the new modules on eggshell}

\begin{table*}[tbh]
\centering{}\caption{\label{tab:setupchoice-2}Settings for ablation study and the results.}
\begin{tabular}{cccccc}
\hline 
\multirow{2}{*}{Ablation setting} & \multirow{2}{*}{Success ($\%$)} & \multirow{2}{*}{Under-drill ($\%$)} & \multicolumn{2}{c}{Over-drill ($\%$)} & \multirow{2}{*}{Average drilling time (min)}\tabularnewline
\cline{4-5}
 &  &  & Model-induced & Human-intervened & \tabularnewline
\hline 
Image only (Baseline) & 80 & 5 & 5 & 10 & 16.8\tabularnewline
Image + Force & 90 & 5 & 5 & 0 & 14.7\tabularnewline
Image + Plane fitting & 85 & \textbf{0} & 0 & 15 & 8.0\tabularnewline
Full & \textbf{95} & 5 & \textbf{0} & \textbf{0} & \textbf{7.1}\tabularnewline
\hline 
\end{tabular}
\end{table*}

To evaluate the impact of the two new modules on system performance:
the force-based completion level recognition and the plane fitting,
we design and implement an ablation study using eggshell drilling
experiments to quantify the contribution of each module by comparing
system performance under different combinations of modules.

\subsubsection{Ablation setting}

Settings for ablation study are summarized in Table~\ref{tab:setupchoice-2},
where the other three groups are to integrate the force-based completion
level recognition module or the plane fitting module into the baseline,
as well as to integrate both modules.

\subsubsection{Metrics}

\begin{figure}[tbh]
\begin{centering}
\includegraphics[width=1\columnwidth]{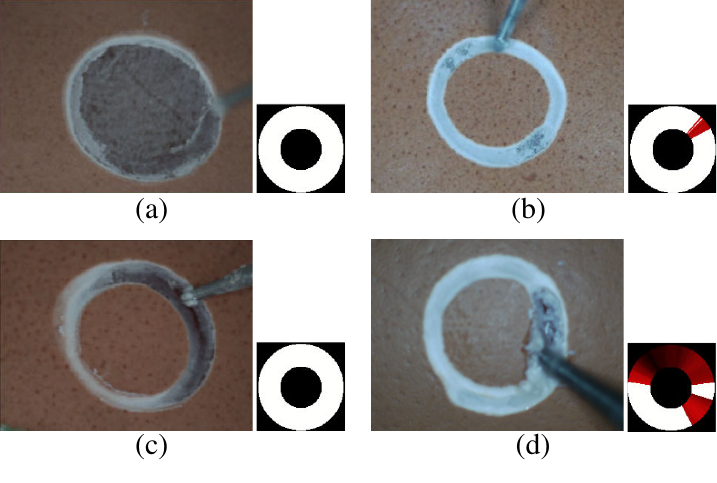}
\par\end{centering}

\caption{\label{fig:result-exp2}Example result images and corresponding progress
bars of (a) a successful case; (b) a failure case of under-drill;
(c) a failure case of model-induced over-drill; and (d) a failure
case of human-intervened over-drill.}
\end{figure}

For each setting, 20 trials are conducted using eggs of random shape,
size, and shell thickness. The drilling procedure ends when the drilling
completion criterion is met (see Section~\ref{subsec:Preparation})
or manually stopped if significant membrane damage is observed. Performance
is evaluated via success rate and average drilling time for successful
cases, with additional failure mode metrics to quantify risks:
\begin{itemize}
\item Success rate: Percentage of trials where the resected eggshell piece
can be manually removed without membrane damage, an example shown
in Fig.~\ref{fig:result-exp2}-(a). Failure occurs if the result
is under-drill or over-drill (model-induced or human-intervened).
\item Under-drill rate: Percentage of trials where the drilling completion
criterion is met, but the eggshell is not manually removable, an example
shown in Fig.~\ref{fig:result-exp2}-(b).
\item Model-induced over-drill rate: Percentage of trials where the drilling
completion criterion is met, but the membrane has been ruptured, an
example shown in Fig.~\ref{fig:result-exp2}-(c).
\item Human-intervened over-drill rate: Percentage of trials where the for
drilling completion criterion is NOT met, but significant membrane
damage is observed and the experiment is compulsory stopped manually,
an example shown in Fig.~\ref{fig:result-exp2}-(d). The human-intervened
over-drill case is the most dangerous case, which could potentially
lead to the death of mice in future experiments with live mice.
\item Average drilling time: Time from drill initiation to autonomous stop,
calculated only for successful trials.
\end{itemize}
These metrics collectively capture trade-offs between precision, speed,
and autonomy, indicating safer and more reliable performance.

\subsubsection{Results and discussion}

\begin{figure}[tbh]
\begin{centering}
\includegraphics[width=1\columnwidth]{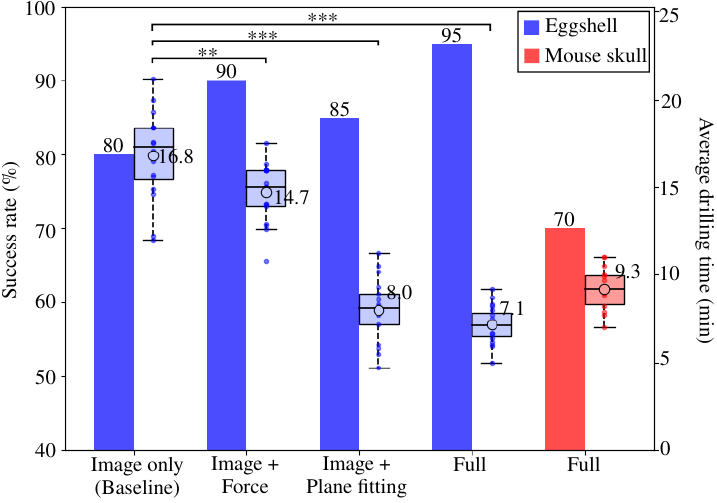}
\par\end{centering}

\caption{\label{fig:comparision of all result}Comparison of the results for
all settings on eggshells (described in Table~\ref{tab:setupchoice-2}),
as well as the result of mouse skull. The bar chart represents the
success rate with different settings while the box plot represents
the average drilling time.}
\end{figure}

Results of all settings are shown in Table~\ref{tab:setupchoice-2},
and the success rates and average drilling time are illustrated in
Fig.~\ref{fig:comparision of all result}.

Compared to the Baseline, integrating the force-based completion level
recognition improved the success rate ($90\,\%$ vs. $80\,\%$) and
reduced the average drilling time (14.7 min vs. 16.8 min) in eggshells.
The introduction of force data allowed information to be updated when
the drilled region was occluded, preventing human-intervened over-drilling
($0\,\%$ vs. $10\,\%$), which is the failure mode we least want
to occur. This reduction also results from the high resolution of
force signal, which resolves ambiguities in drilling completion level
recognition solely based on images and allows more accurate recognition
when near penetration. However, $5\,\%$ model-induced over-drilling
persisted due to long drilling time, where the drill bit ruptures
the membrane surface despite retaining its $z$-position at points
reaching a drilling completion level of 1 when the drill is trying
to drill the other points that are not. Despite this, the total over-drilling
risk decreased from $15\,\%$ to $5\,\%$, demonstrating the benefit
of force-based completion level recognition.

Integrating the plane fitting significantly improved the drilling
speed (8.0 min vs. 16.8 min) and slightly improved the success rate
($85\,\%$ vs. $80\,\%$) in eggshells, compared to the Baseline.
This suggests that the long drilling time and failures caused by varying
egg initial poses in our former research were overcome by the plane
fitting. However, the accelerated drilling speed occasionally exceeds
the system's ability to dynamically update drilling completion levels,
leading to a higher rate of human-intervened over-drill ($15\,\%$
vs. $10\,\%$) and increasing risks.

Integration of the two new modules achieved the highest success rate
($95\,\%$) and fastest drilling (7.1 min), while eliminating all
over-drilling risks ($0\,\%$ model-induced and $0\,\%$ human-intervened
over-drill). Noting that the only failure was caused by under-drill,
meaning that the drilling completion criterion was met so the drilling
stopped automatically, but the eggshell inside the drilling area could
not be removed by tweezers manually. This suggests that in future
work the drilling completion criterion needs to be adjusted. For example,
an iterative learning system could be implemented to refine the threshold
that is currently constant. Alternatively, a non-rotating drill bit
could be controlled to automatically press the eggshell within the
drilling area and observe its displacement to determine whether it
is detachable, mimicking the same inspection procedure performed by
human operators in practical mice cranial window creation. Additionally,
a t-test was conducted to evaluate the differences in performance
between the integration groups and the Baseline and the p-values are
shown in Fig.~\ref{fig:comparision of all result}, demonstrating
that integrating either of the new modules, or both of the modules
can lead to statistically significant improvements in system performance.

\begin{figure*}[tbh]
\begin{centering}
\includegraphics[width=1\textwidth]{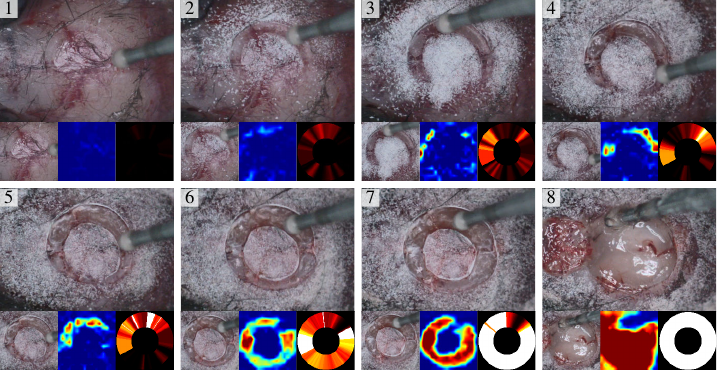}
\par\end{centering}

\caption{\label{fig:drillingsnapshots}A series of snapshots of a case of successful
drilling on a \emph{postmortem} mouse skull from the start of drilling
to the removal of resected circular shell piece. Original image, cropped
image, heatmap (generated by image processing), and drilling progress
bar (generated based on the estimation result of both image and force)
are shown for each snapshot.}
\end{figure*}

\begin{table*}[tbh]
\begin{centering}
\caption{\label{tab:comparisonwithrelated work}Comparison of the proposed
work in contrast with existing literature.}
\par\end{centering}
\centering{}%
\begin{tabular}{|c|c|c||c|c|c|c|c|c|c|}
\hline 
Object & Metrics & This work & \cite{Zhao2023} & \cite{Pak2015} & \cite{Ghanbari2019} & \cite{Jeong2013} & \cite{hasegawa2023} & \cite{Jia2023} & \cite{Navabi25}\tabularnewline
\hline 
\hline 
\multirow{2}{*}{Egg} & Success rate ($\%$) & \multirow{1}{*}{$95$ ($n=20$)} & $80$ ($n=20$) & N.A. & N.A. & N.A. & $100$ ($n=2$) & N.A. & N.A.\tabularnewline
\cline{2-10}
 & Average time (min) & 7.1 & 16.8 & N.A. & N.A. & N.A. & N.A. & N.A. & N.A.\tabularnewline
\hline 
\multirow{2}{*}{Mice} & Success rate ($\%$) & $70$ ($n=20$) & N.A. & N.A. & N.A. & N.A. & N.A. & N.A. & N.A.\tabularnewline
\cline{2-10}
 & Average time (min) & 9.3 & N.A. & $<15$ ($n=15$) & 10-15 ($n=6$) & N.A. & N.A. & N.A. & $<5$ ($n=2$)\tabularnewline
\hline 
\end{tabular}
\end{table*}

\subsection{Experiment 3: validation experiment on postmortem mouse skull}

In Experiment 2, we showed that with the integration of new modules
the autonomous robotic drilling system performed well on eggshells,
however, it is still necessary to verify the feasibility on mouse
skull drilling. 20 trials on 20 different\emph{ postmortem} mice are
conducted. Similar to the eggshell experiment, the resected circular
mouse skull patch being removable manually with tweezers defines successes
and failures.

\subsubsection*{Results and discussion}

In the experiment, 14 out of 20 cases were successful, resulting in
a success rate of $70\,\%$. The average required time for successful
drilling was $9.3$ min. The result of \emph{postmortem} mouse skull
drilling is also shown in Fig.~\ref{fig:comparision of all result},
with a representative successful case illustrated in Fig.~\ref{fig:drillingsnapshots}.
In the case, the drilling procedure stopped automatically when it
was judged as complete and the circular patch was removed perfectly.
We can conclude that the system effective for mouse skull drilling
with a considerable success rate and drilling speed.

Five of the six failure cases were human-intervened over-drill because
of incorrect completion level recognition, and the remaining one was
under-drill. Compared with the results of eggshell drilling, the completion
level recognition is considered less accurate for mouse skulls due
to more complex images and textures. For image-based recognition,
the color changes in mouse skulls are subtler and influenced by factors
like individual variation, skull humidity, and the time between euthanasia
and the experiment. For example, moist skulls shortly after euthanasia
show minimal color change during drilling, while drier skulls appear
whiter as drilling progresses. These variations complicate recognition.
Force-based recognition is also affected by skull moistness, which
alters stiffness, and by fluid exudation during drilling. Bleeding
in some trials changed both stiffness and color, further reducing
accuracy. Additionally, reflected light from wet skull surfaces negatively
impacts recognition. Applying more data augmentation methods to the
mouse skull dataset (e.g., random bleeding, random light reflection),
generating photo-realistic synthetic data in a more complex environment,
and adding random noise to the force training dataset could be potential
future work to further enhance the robustness of our system.

\subsection{Conclusion and discussion}

The results of the evaluation experiments with different settings
(described in Table~\ref{tab:setupchoice-2}) on eggshell and mouse
skulls are shown in Fig.~\ref{fig:comparision of all result}, which
demonstrate the effectiveness and robustness of our autonomous robotic
drilling system. More complex models of neural network for both image-based
and force-based recognition, which are more robust to changes in the
color or moistness change, can help to some extent but processing
speed can suffer. Additional sources of information, such as audio
or vibration, can be added in future work to improve robustness.

We compared our results with those of other related studies on creating
a cranial window of similar size ($d=8\,\text{mm}$), as shown in
Table~\ref{tab:comparisonwithrelated work}. Note that most works
did not conduct repeat experiments to evaluate their system with a
success rate and only provided a vague drilling time, which results
in the difficulties in compare the robustness and speed. We are the
only research to evaluate the system with a general metrics to demonstrate
the performance.

\section{Conclusion}

In this paper, we propose an autonomous robotic drilling system for
mice cranial window creation. To achieve this, a trajectory planner
that can adapt to the drilled object and prevent overshoot is imposed,
and a plane fitting method is added to solve the problem caused by
long drilling time. In order to adjust the generated trajectory in
real-time, we apply neural networks to recognize the drilling completion
level based on image and force data obtained during the drilling procedure.
In the experiments, we show that the proposed system is able to achieve
a considerable success rate and speed in autonomous eggshell drilling
and demonstrate the possibility of the system on autonomous mice cranial
window creation.

Different from existing studies, which either focus on local penetration
detection, rely on pre-operative global information, or employ specialized
sensors or simulations, our system integrates perceived contact signals
and global information to successfully achieve a real-time, adaptive,
and generalizable drilling trajectory generation and planning method
without pre-processing. Additionally, we provide experimental metrics,
including drilling speed and success rate, establishing a standard
for comparing related studies.

Our ultimate goal is the autonomous cranial window creation on anesthetized
live mice which require the mice to be alive and remain alive after
the procedure. This is a considerable challenge in aspects that are
being investigated in parallel to this work, such as surgical conditions,
anesthesia, sterilization, ethical board approval, and other aspects.
In this work we present a clear step towards this goal. To further
approach this ultimate goal, future work will include enhancing the
robustness of the completion level recognition block by fine-tuning
the neural network, adopting a more accurate and objective drilling
completion criterion, and improving the success rate of drilling by
incorporating additional modalities such as audio. Furthermore, given
that many surgeries, including orthognathic surgery and arthroplasty,
face similar challenges related to target uncertainty in surface and
thickness, we aim to extend the application of our autonomous robotic
drilling system to other medical fields.

\bibliographystyle{IEEEtran}
\bibliography{ICRA2022,IEEEabrv,IEEEexample}

\begin{thebibliography}{10}
\providecommand{\url}[1]{#1}
\csname url@samestyle\endcsname
\providecommand{\newblock}{\relax}
\providecommand{\bibinfo}[2]{#2}
\providecommand{\BIBentrySTDinterwordspacing}{\spaceskip=0pt\relax}
\providecommand{\BIBentryALTinterwordstretchfactor}{4}
\providecommand{\BIBentryALTinterwordspacing}{\spaceskip=\fontdimen2\font plus
\BIBentryALTinterwordstretchfactor\fontdimen3\font minus
  \fontdimen4\font\relax}
\providecommand{\BIBforeignlanguage}[2]{{%
\expandafter\ifx\csname l@#1\endcsname\relax
\typeout{** WARNING: IEEEtran.bst: No hyphenation pattern has been}%
\typeout{** loaded for the language `#1'. Using the pattern for}%
\typeout{** the default language instead.}%
\else
\language=\csname l@#1\endcsname
\fi
#2}}
\providecommand{\BIBdecl}{\relax}
\BIBdecl

\bibitem{Marinho2024}
M.~M. Marinho, J.~J. Quiroz-Oma{\~n}a, and K.~Harada, ``A multiarm robotic
  platform for scientific exploration: Its design, digital twins, and
  validation,'' \emph{IEEE Robotics \& Automation Magazine}, pp. 2--12, 2024.

\bibitem{Koike2019}
H.~Koike, K.~Iwasawa, R.~Ouchi, M.~Maezawa, K.~Giesbrecht, N.~Saiki,
  A.~Ferguson, M.~Kimura, W.~L. Thompson, J.~M. Wells, A.~M. Zorn, and
  T.~Takebe, ``Modelling human hepato-biliary-pancreatic organogenesis from the
  foregut{\textendash}midgut boundary,'' \emph{Nature}, vol. 574, no. 7776, pp.
  112--116, sep 2019.

\bibitem{Holtmaat2009}
A.~Holtmaat, T.~Bonhoeffer, D.~K. Chow, J.~Chuckowree, V.~De~Paola, S.~B.
  Hofer, M.~Hübener, T.~Keck, G.~Knott, W.-C.~A. Lee, R.~Mostany, T.~D.
  Mrsic-Flogel, E.~Nedivi, C.~Portera-Cailliau, K.~Svoboda, J.~T. Trachtenberg,
  and L.~Wilbrecht, ``Long-term, high-resolution imaging in the mouse neocortex
  through a chronic cranial window,'' \emph{Nature Protocols}, vol.~4, no.~8,
  pp. 1128--1144, Jul. 2009.

\bibitem{Goldey2014}
G.~J. Goldey, D.~K. Roumis, L.~L. Glickfeld, A.~M. Kerlin, R.~C. Reid,
  V.~Bonin, D.~P. Schafer, and M.~L. Andermann, ``Removable cranial windows for
  long-term imaging in awake mice,'' \emph{Nature Protocols}, vol.~9, no.~11,
  pp. 2515--2538, Oct. 2014.

\bibitem{Drew2010}
P.~J. Drew, A.~Y. Shih, J.~D. Driscoll, P.~M. Knutsen, P.~Blinder, D.~Davalos,
  K.~Akassoglou, P.~S. Tsai, and D.~Kleinfeld, ``Chronic optical access through
  a polished and reinforced thinned skull,'' \emph{Nature Methods}, vol.~7,
  no.~12, pp. 981--984, Oct. 2010.

\bibitem{Yang2012}
G.~Yang and W.~Gan, ``Sleep contributes to dendritic spine formation and
  elimination in the developing mouse somatosensory cortex,''
  \emph{Developmental Neurobiology}, vol.~72, no.~11, pp. 1391--1398, Jul.
  2012.

\bibitem{Kim2016}
T.~H. Kim, Y.~Zhang, J.~Lecoq, J.~C. Jung, J.~Li, H.~Zeng, C.~M. Niell, and
  M.~J. Schnitzer, ``Long-term optical access to an estimated one million
  neurons in the live mouse cortex,'' \emph{Cell Reports}, vol.~17, no.~12, pp.
  3385--3394, Dec. 2016.

\bibitem{Liu2024}
\BIBentryALTinterwordspacing
Y.~Liu, D.~Song, G.~Zhang, Q.~Bu, Y.~Dong, C.~Hu, and C.~Shi, ``A novel
  electromagnetic driving system for 5-dof manipulation in intraocular
  microsurgery,'' \emph{Cyborg and Bionic Systems}, vol.~5, p. 0083, 2024.
  [Online]. Available:
  \url{https://spj.science.org/doi/abs/10.34133/cbsystems.0083}
\BIBentrySTDinterwordspacing

\bibitem{Wang2024}
\BIBentryALTinterwordspacing
T.~Wang, T.~Jin, W.~Lin, Y.~Lin, H.~Liu, T.~Yue, Y.~Tian, L.~Li, Q.~Zhang, and
  C.~Lee, ``Multimodal sensors enabled autonomous soft robotic system with
  self-adaptive manipulation,'' \emph{ACS Nano}, vol.~18, no.~14, pp.
  9980--9996, 2024, pMID: 38387068. [Online]. Available:
  \url{https://doi.org/10.1021/acsnano.3c11281}
\BIBentrySTDinterwordspacing

\bibitem{Pohl2011}
B.~M. Pohl, A.~Schumacher, and U.~G. Hofmann, ``Towards an automated, minimal
  invasive, precision craniotomy on small animals,'' in \emph{2011 5th
  International {IEEE}/{EMBS} Conference on Neural Engineering}.\hskip 1em plus
  0.5em minus 0.4em\relax {IEEE}, apr 2011.

\bibitem{andreoli2018egg}
L.~Andreoli, H.~Simpl{\'\i}cio, and E.~Morya, ``Egg model training protocol for
  stereotaxic neurosurgery and microelectrode implantation,'' \emph{World
  Neurosurgery}, vol. 111, pp. 243--250, 2018.

\bibitem{okuda2010training}
T.~Okuda, K.~Kataoka, and A.~Kato, ``Training in endoscopic endonasal
  transsphenoidal surgery using a skull model and eggs,'' \emph{Acta
  neurochirurgica}, vol. 152, pp. 1801--1804, 2010.

\bibitem{Zhao2023}
E.~Zhao, M.~M. Marinho, and K.~Harada, ``Autonomous robotic drilling system for
  mice cranial window creation: An evaluation with an egg model,'' in
  \emph{2023 IEEE/RSJ International Conference on Intelligent Robots and
  Systems (IROS)}.\hskip 1em plus 0.5em minus 0.4em\relax IEEE, 2023, pp.
  4592--4599.

\bibitem{Pak2015}
N.~Pak, J.~H. Siegle, J.~P. Kinney, D.~J. Denman, T.~J. Blanche, and E.~S.
  Boyden, ``Closed-loop, ultraprecise, automated craniotomies,'' \emph{Journal
  of Neurophysiology}, vol. 113, no.~10, pp. 3943--3953, jun 2015.

\bibitem{Ghanbari2019}
L.~Ghanbari, M.~L. Rynes, J.~Hu, D.~S. Schulman, G.~W. Johnson, M.~Laroque,
  G.~M. Shull, and S.~B. Kodandaramaiah, ``Craniobot: A computer numerical
  controlled robot for cranial microsurgeries,'' \emph{Scientific Reports},
  vol.~9, no.~1, jan 2019.

\bibitem{Jeong2013}
D.~C. Jeong, P.~S. Tsai, and D.~Kleinfeld, ``All-optical osteotomy to create
  windows for transcranial imaging in mice,'' \emph{Optics Express}, vol.~21,
  no.~20, p. 23160, Sep. 2013.

\bibitem{hasegawa2023}
S.~Hasegawa, K.~Okada, and M.~Inaba, ``\BIBforeignlanguage{ja}{Vibration
  {Recognition} {Learning} {System} using {Force} and {Sound} {Sensing}
  integratively for {Mouse} {Cranium} {Cutting} in {Cranial} {Window}
  {Surgery}},'' in \emph{\BIBforeignlanguage{ja}{2023 {JSME} {Conference} on
  {Robotics} and {Mechatronics}}}, Nagoya, Japan, 2023.

\bibitem{Jia2023}
Y.~Jia, P.~M. Uriguen~Eljuri, and T.~Taniguchi, ``A bayesian reinforcement
  learning method for periodic robotic control under significant uncertainty,''
  in \emph{2023 IEEE/RSJ International Conference on Intelligent Robots and
  Systems (IROS)}.\hskip 1em plus 0.5em minus 0.4em\relax IEEE, Oct. 2023.

\bibitem{Navabi25}
\BIBentryALTinterwordspacing
Z.~S. Navabi, R.~Peters, B.~Gulner, A.~Cherkkil, E.~Ko, F.~Dadashi, J.~O.
  Brien, M.~Feldkamp, and S.~B. Kodandaramaiah, ``Computer vision-guided rapid
  and precise automated cranial microsurgeries in mice,'' \emph{Science
  Advances}, vol.~11, no.~15, p. eadt9693, 2025. [Online]. Available:
  \url{https://www.science.org/doi/abs/10.1126/sciadv.adt9693}
\BIBentrySTDinterwordspacing

\bibitem{Dvornik2017}
N.~Dvornik, K.~Shmelkov, J.~Mairal, and C.~Schmid, ``Blitznet: A real-time deep
  network for scene understanding,'' in \emph{Proceedings of the IEEE
  international conference on computer vision}, 2017, pp. 4154--4162.

\bibitem{Fu2017}
C.-Y. Fu, W.~Liu, A.~Ranga, A.~Tyagi, and A.~C. Berg, ``Dssd : Deconvolutional
  single shot detector,'' 2017.

\bibitem{hochreiter1997long}
S.~Hochreiter and J.~Schmidhuber, ``Long short-term memory,'' \emph{Neural
  computation}, vol.~9, no.~8, pp. 1735--1780, 1997.

\bibitem{He2019}
C.~He, N.~Patel, A.~Ebrahimi, M.~Kobilarov, and I.~Iordachita, ``Preliminary
  study of an rnn-based active interventional robotic system (airs) in retinal
  microsurgery,'' \emph{International Journal of Computer Assisted Radiology
  and Surgery}, vol.~14, no.~6, pp. 945--954, Mar. 2019.

\bibitem{Li2023}
R.~Li, Z.~Dong, J.-M. Wu, C.~J. Xue, and N.~Guan, ``Modeling and property
  analysis of the message synchronization policy in ros,'' in \emph{2023 IEEE
  International Conference on Mobility, Operations, Services and Technologies
  (MOST)}.\hskip 1em plus 0.5em minus 0.4em\relax IEEE, May 2023.

\bibitem{kruger2003constrained}
C.~Kruger, ``Constrained cubic spline interpolation,'' \emph{Chemical
  Engineering Applications}, vol.~1, no.~1, 2003.

\bibitem{Kumaar2021}
S.~Kumaar, Y.~Lyu, F.~Nex, and M.~Y. Yang, ``Cabinet: Efficient context
  aggregation network for low-latency semantic segmentation,'' in \emph{2021
  IEEE International Conference on Robotics and Automation (ICRA)}.\hskip 1em
  plus 0.5em minus 0.4em\relax IEEE, May 2021, pp. 13\,517--13\,524.

\bibitem{Yu2021}
C.~Yu, C.~Gao, J.~Wang, G.~Yu, C.~Shen, and N.~Sang, ``Bisenet v2: Bilateral
  network with guided aggregation for real-time semantic segmentation,''
  \emph{International Journal of Computer Vision}, vol. 129, no.~11, pp.
  3051--3068, Sep. 2021.

\bibitem{Peng2022}
J.~Peng, Y.~Liu, S.~Tang, Y.~Hao, L.~Chu, G.~Chen, Z.~Wu, Z.~Chen, Z.~Yu,
  Y.~Du, Q.~Dang, B.~Lai, Q.~Liu, X.~Hu, D.~Yu, and Y.~Ma, ``Pp-liteseg: A
  superior real-time semantic segmentation model,'' 2022.

\bibitem{Li2020}
X.~Li, A.~You, Z.~Zhu, H.~Zhao, M.~Yang, K.~Yang, S.~Tan, and Y.~Tong,
  \emph{Semantic Flow for Fast and Accurate Scene Parsing}.\hskip 1em plus
  0.5em minus 0.4em\relax Springer International Publishing, 2020, pp.
  775--793.

\bibitem{Pan2023}
H.~Pan, Y.~Hong, W.~Sun, and Y.~Jia, ``Deep dual-resolution networks for
  real-time and accurate semantic segmentation of traffic scenes,'' \emph{IEEE
  Transactions on Intelligent Transportation Systems}, vol.~24, no.~3, pp.
  3448--3460, Mar. 2023.

\bibitem{Xu_2023_CVPR}
J.~Xu, Z.~Xiong, and S.~P. Bhattacharyya, ``Pidnet: A real-time semantic
  segmentation network inspired by pid controllers,'' in \emph{Proceedings of
  the IEEE/CVF Conference on Computer Vision and Pattern Recognition (CVPR)},
  June 2023, pp. 19\,529--19\,539.

\bibitem{adorno2021dqrobotics}
B.~V. Adorno and M.~M. Marinho, ``Dq robotics: A library for robot modeling and
  control,'' \emph{IEEE Robotics \& Automation Magazine}, vol.~28, no.~3, pp.
  102--116, 2020.

\end{thebibliography}

\begin{IEEEbiography}[{\includegraphics[width=1in,height=1.25in]{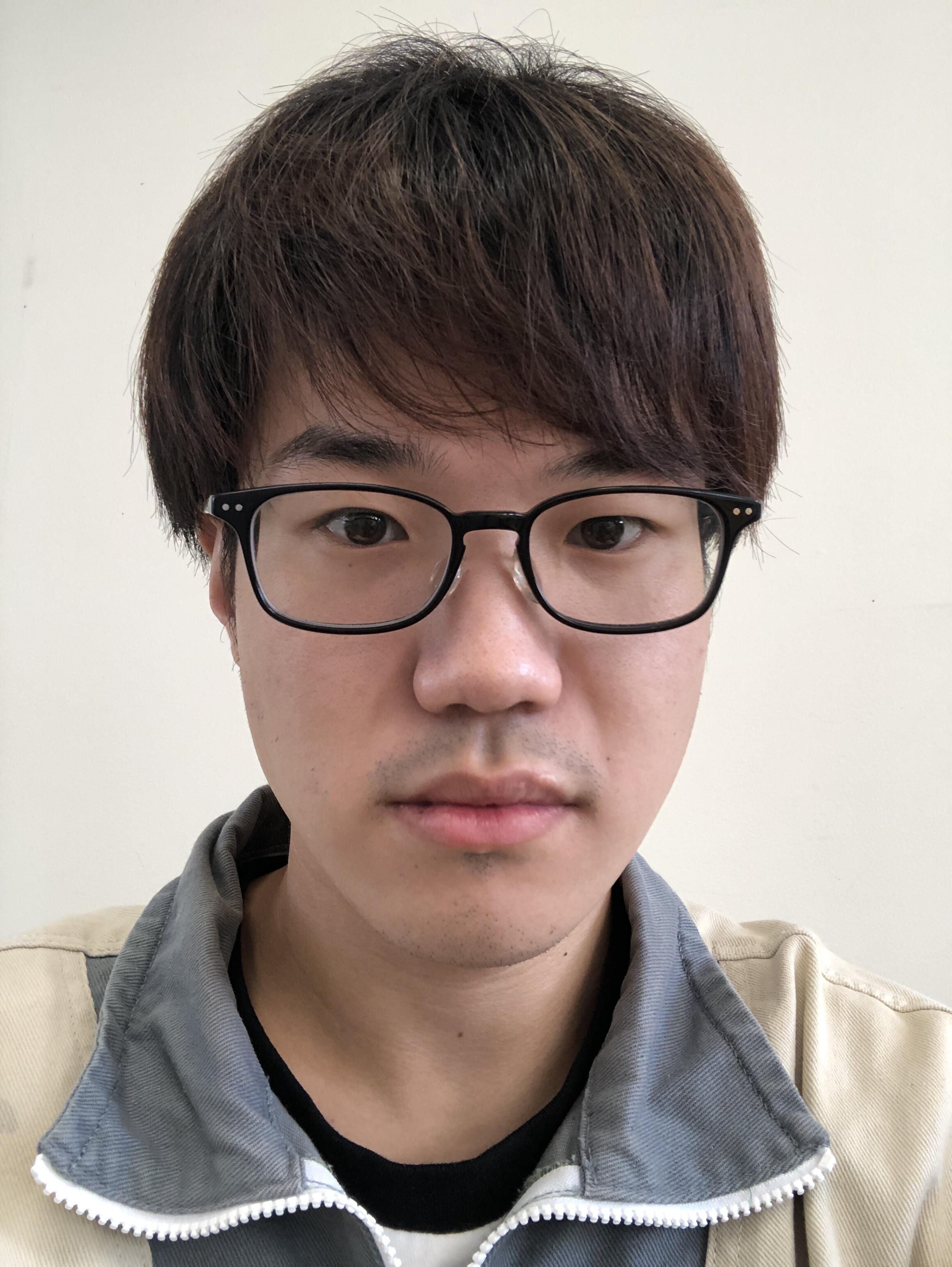}}]{Enduo Zhao}
 is a postdoctoral researcher at the School of Biomedical Engineering,
Tsinghua University, Beijing, China. He received a BS (2018) in mechanical
engineering from Tsinghua University, Beijing, China, a MSc (2020)
and a Ph.D. (2025) in mechanical engineering from the University of
Tokyo, Tokyo, Japan. His research interests include automation, surgical
robotics, and medical image processing.
\end{IEEEbiography}

\begin{IEEEbiography}[{\includegraphics[width=1in,height=1.25in]{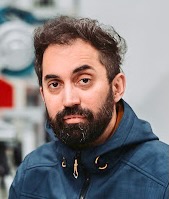}}]{Murilo M. Marinho}
 (GS'13\textendash M'18\textendash SM'24) is a visiting researcher
with the Graduate School of Medicine, the University of Tokyo, Tokyo,
Japan. He is also a Lecturer in Robotics with the University of Manchester,
Manchester, UK. Until late 2023, he was an Assistant Professor with
the University of Tokyo. He received a BS (2012) in mechatronics engineering
and a MSc (2014) in electronic systems and automation engineering
from the University of Brasilia, Brasilia, Brazil, and a Ph.D. (2018)
in mechanical engineering from the University of Tokyo, Tokyo, Japan.
His research interests include robotics applied to constrained workspaces
and image-based automation.
\end{IEEEbiography}

\begin{IEEEbiography}[{\includegraphics[width=1in,height=1.25in]{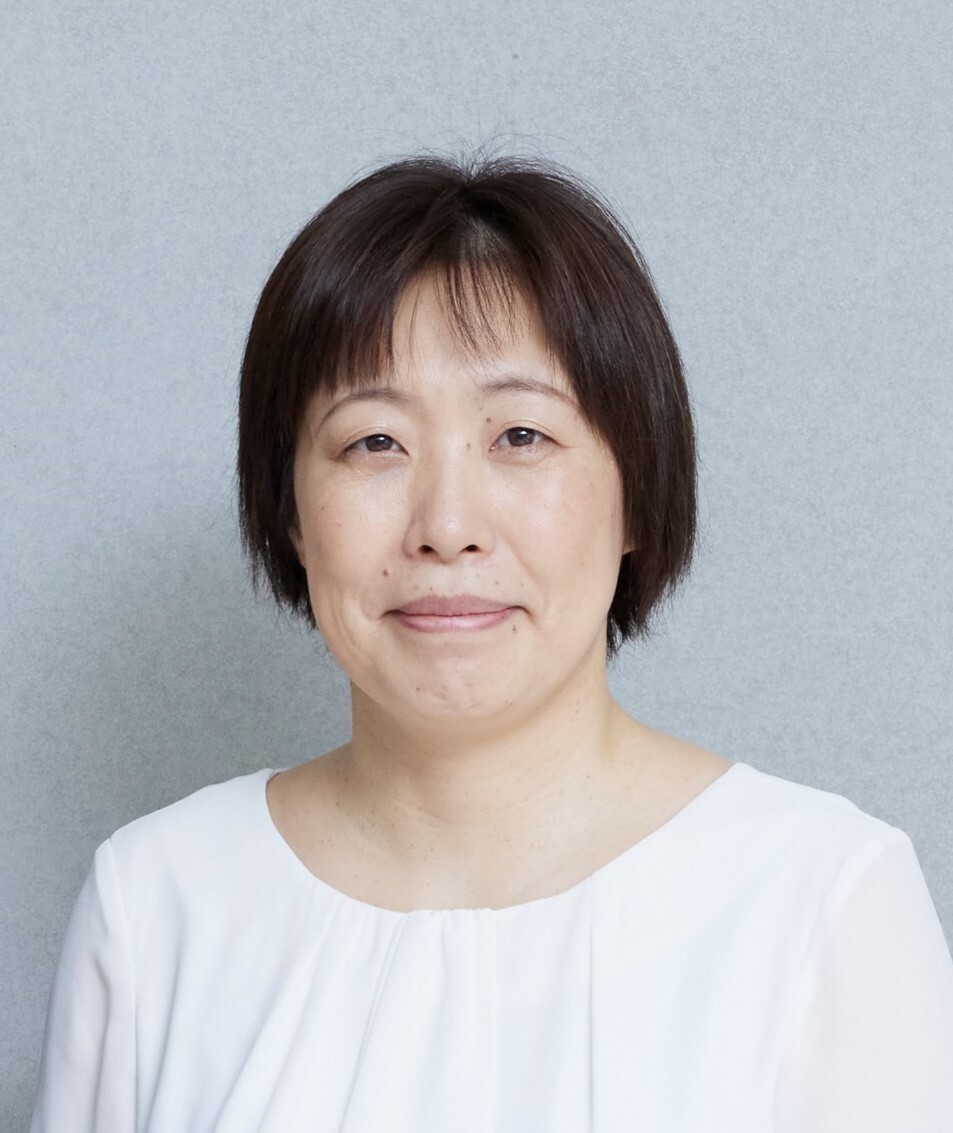}}]{Kanako Harada}
 (M'07) Kanako Harada is a Professor at the Center for Disease Biology
and Integrative Medicine (CDBIM) within the Graduate School of Medicine
at The University of Tokyo, Japan. She also holds positions in the
Department of Mechanical Engineering and the Department of Bioengineering
within the Graduate School of Engineering. She earned her M.Sc. in
Engineering from The University of Tokyo in 2001 and received her
Ph.D. in Engineering from Waseda University in 2007. Before joining
The University of Tokyo, she held positions at Hitachi Ltd., the Japan
Association for the Advancement of Medical Equipment, and Scuola Superiore
Sant\textquoteright Anna in Italy. Her research interests include
surgical robotic systems, robotic automation for biomedical applications
and regulatory science.

\end{IEEEbiography}

\end{document}